\documentclass[10pt,twocolumn,letterpaper]{article}

\usepackage[pagenumbers]{cvpr} % To force page numbers, e.g. for an arXiv version
\makeatother  
\definecolor{cvprblue}{rgb}{0.21,0.49,0.74}
\usepackage[pagebackref,breaklinks,colorlinks,allcolors=cvprblue]{hyperref}
\usepackage[ruled,linesnumbered]{algorithm2e}
\usepackage{multirow}
\usepackage{amsmath, amssymb}  % 数学命令
\usepackage{graphicx}           % \includegraphics
\usepackage{hyperref}           % \url, \href
\usepackage{natbib}             % \citep, \citet
\def\paperID{8637} % *** Enter the Paper ID here
\def\confName{CVPR}
\def\confYear{2026}

\title{TinySR: Pruning Diffusion for Real-World Image Super-Resolution}

\def\spaces{~~~~~~}
\author{
Linwei Dong\textsuperscript{1}\spaces{}
Qingnan Fan\textsuperscript{2}\spaces{}
Yuhang Yu\textsuperscript{2}\spaces{} 
Qi Zhang\textsuperscript{2}\spaces{} \\
Jinwei Chen\textsuperscript{2}\spaces{} 
Yawei Luo\textsuperscript{1}$^{\dagger}$\spaces{}
Changqing Zou\textsuperscript{1, 3}\\
\textsuperscript{1}Zhejiang University\spaces{}
\textsuperscript{2}Vivo Mobile Communication Co. Ltd\spaces{} 
\textsuperscript{3}Zhejiang Lab
}

\begin{document}
\maketitle
\begin{abstract}
Real-world image super-resolution (Real-ISR) focuses on recovering high-quality images from low-resolution inputs that suffer from complex degradations like noise, blur, and compression. Recently, diffusion models (DMs) have shown great potential in this area by leveraging strong generative priors to restore fine details. However, their iterative denoising process incurs high computational overhead, posing challenges for real-time applications. Although one-step distillation methods, such as OSEDiff and TSD-SR, offer faster inference, they remain fundamentally constrained by their large, over-parameterized model architectures. In this work, we present \textbf{TinySR}, a compact yet effective diffusion model specifically designed for Real-ISR that achieves real-time performance while maintaining perceptual quality. We introduce \textbf{Dynamic Inter-block Activation} and \textbf{Expansion-Corrosion Strategy} to facilitate more effective decision-making in depth pruning. We achieve VAE compression through channel pruning, attention blocks removal and lightweight \textit{SepConv}. We eliminate time- and prompt-related modules and perform pre-caching techniques to further speed up the model. \textbf{TinySR} significantly reduces computational cost and model size, achieving up to \textbf{5.68$\times$} speedup and \textbf{83\% }parameter reduction compared to its teacher TSD-SR, while still providing high quality results.
Our code is released at \href{https://github.com/Microtreei/TinySR}{\textcolor{pink}{https://github.com/Microtreei/TinySR}}.
\end{abstract}

\section{Introduction}
\label{sec:intro}
Real-world image super-resolution (Real-ISR) \cite{zhang2021designing, wang2021real} aims to reconstruct high-fidelity images from low-quality observations corrupted by compound degradations including noise contamination, nonlinear blur, and compression artifacts.
Recently, diffusion models (DMs) have demonstrated significant promise for Real-ISR by leveraging their powerful priors to effectively address complex degradation patterns while recovering realistic textures and details \cite{ho2020denoising, nichol2021improved}.
Their impressive performance over GAN-based Real-ISR methods \cite{liang2022details} has driven their widespread adoption in practical downstream applications.
However, for resource-constrained environments, the iterative sampling nature and high computational demands of DMs fundamentally limit their deployment \cite{chen2025snapgen}.

Significant efforts in reducing diffusion model sampling steps have dramatically improved the inference latency of DM-based Real-ISR approaches.
Recent advancements in efficient Real-ISR models, such as OSEDiff \cite{wu2024one} and TSD-SR \cite{dong2024tsd} seek to condense the denoising process into a single step by carefully designed distillation while preserving high-quality restoration performance. 
However, these methods still depend on large pre-trained models, which require significant computational resources and footprint, posing challenges for real-time applications and edge-device deployment. 
There is a critical demand for more efficient and compact models that can be readily achieved while maintaining competitive performance.
\begin{figure}[!t]
  \centering
    \includegraphics[width=\linewidth]{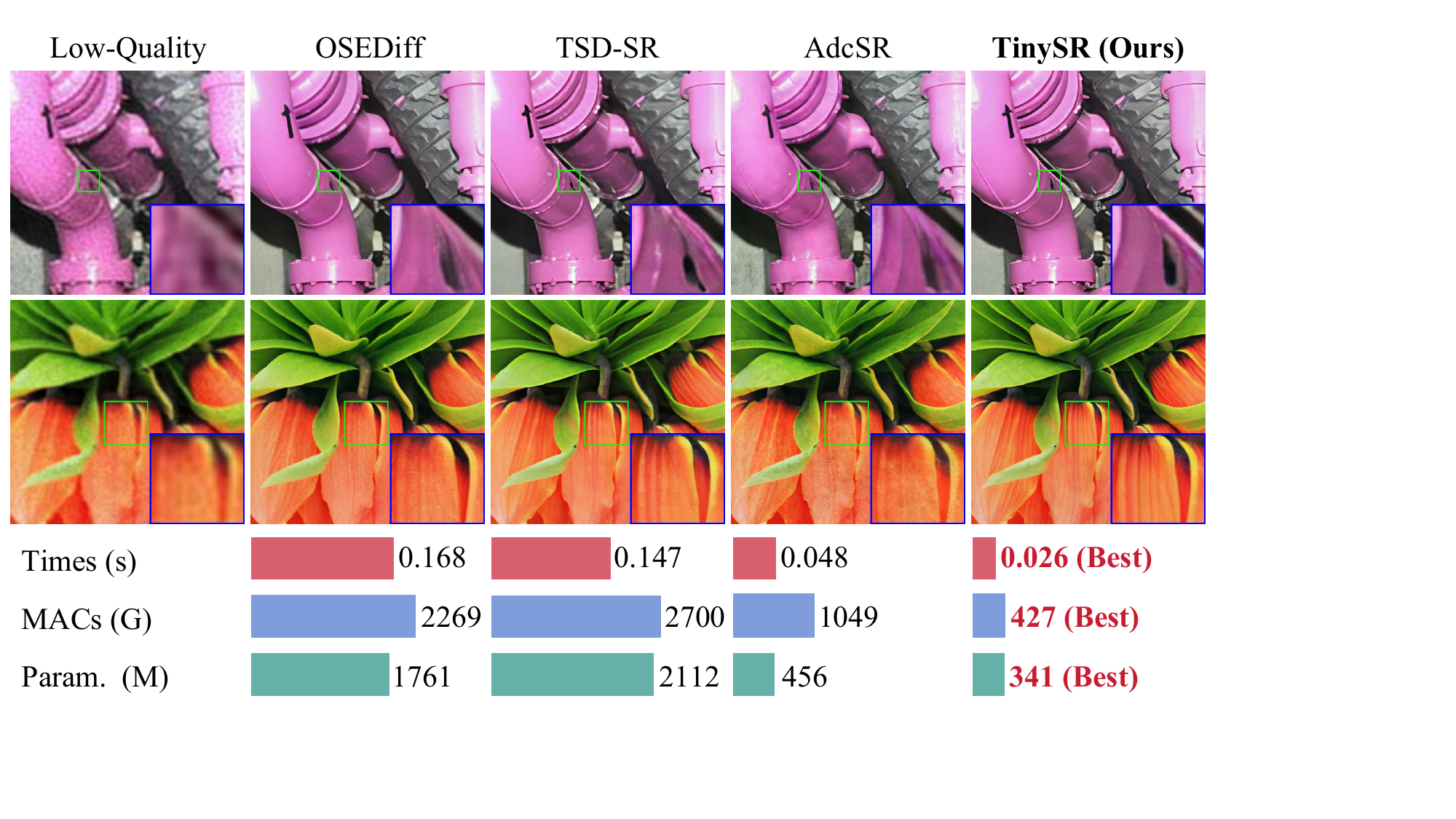}
   \caption{A comprehensive comparison of recent Real-ISR models in terms of visual quality, inference time, computational cost (MACs), and parameter count, highlighting the superior efficiency and performance of our proposed method.
   }
   \label{fig:teaser}
\end{figure}

Diffusion compression primarily involves several key techniques, including operator optimization \cite{dao2022flashattention, teng2024dim}, precision quantization \cite{he2023ptqd, li2023q}, width pruning \cite{wang2023patch, castells2024ld, chen2025snapgen}, and depth pruning \cite{flux1-lite, fang2024tinyfusion, kim2024bk, li2023snapfusion}.
Depth pruning serves as a simple but effective technique, favored for its linear acceleration and straightforward implementation.
The conventional paradigm for depth pruning is dependent upon empirical \cite{kim2024bk} or importance-based metrics \cite{men2024shortgpt, han2015learning} to guide layer selection, yet it overlooks the recoverability of the model's performance on its target task after being pruned.
Instead of relying on static importance scores, probability-based mask learning aims to iteratively refine this sampling distribution, such that layers with greater performance recoverability are more likely to be sampled.
However, in ultra-deep networks, mask learning confronts a combinatorial explosion in the search space, which results in prohibitive optimization complexity and slow convergency \cite{fang2024tinyfusion}.

Beyond deep architecture limitation, DMs-based Real-ISR models exhibit two other computational bottlenecks: high overhead from VAE (Variational Auto-Encoder) \cite{kingma2013auto} and inefficiencies from redundant condition modules during super-resolution process. 
AdcSR \cite{chen2024adversarial} eliminates the VAE encoder by employing a \textit{PixelUnshuffle} operation \cite{shi2016real}.
However, fine-grained channel pruning is required in denoising networks to align channel dimensions, which significantly increases overall complexity and coupling between its components.
Regarding prompts and time embeddings, recent studies \cite{dong2024tsd, chen2024adversarial} indicate that these conditions contribute minimally to the one-step Real-ISR model, suggesting that a more efficient model can be achieved by eliminating these inputs and related modules.

\begin{figure}[!t]
  \centering
    \includegraphics[width=\linewidth]{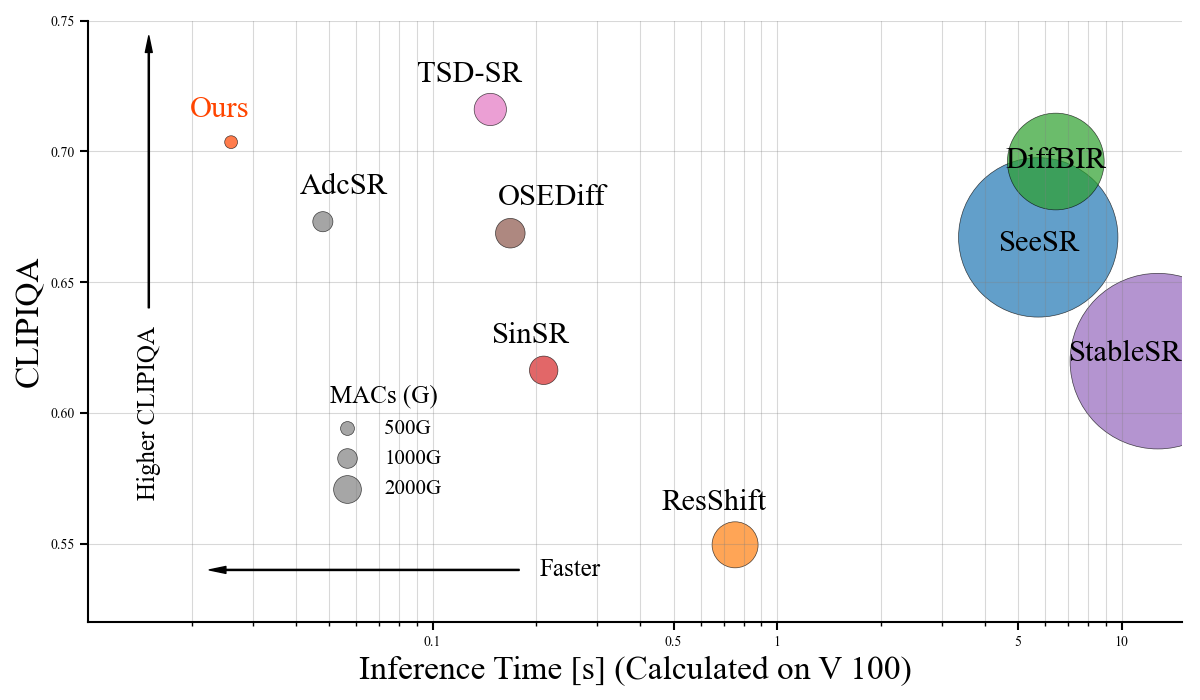}
   \caption{Performance and efficiency comparison among DMs-based Real-ISR methods on an NVIDIA V100
GPU. All metrics are evaluated on the RealSR benchmark.
\textbf{TinySR} achieves the fastest inference, lightest computation (low MACs) and commendable performance (high CLIPIQA).
   }
   \label{fig:bubble}
\end{figure}

Based on these analyses, we propose {\bf TinySR}, a compact yet effective DMs-based Real-SR model that eliminates computational redundancies in TSD-SR while maintaining restoration quality.
Following mask learning, we propose identifying candidate layers that exhibit high-performance recoverability.
We partition the network into non-overlapping blocks and introduce sets of learnable probabilities $p(m)$ for each blcok to constrain search space.
We propose \textbf{Dynamic Inter-block Activation}, a method that leverages the learnable probability $q(t)$ for soft boundary exploration, and introduce a novel \textbf{Expansion-Corrosion Strategy} to determine the optimal pruning scheme.
These proposed techniques involve a strategic trade-off between optimization complexity and exploratory potential.
Furthermore, we propose several strategies to further compress the Real-ISR model.
To lighten the VAE, we perform channel-wise pruning, remove its computationally intensive attention modules, and replace its standard convolutions with depthwise separable convolutions \cite{howard2017mobilenets}.
We eliminate time- and prompt-related modules to further enhance computational efficiency.
Extensive experiments on standard Real-ISR benchmarks demonstrate that our method is up to \textbf{5.68$\times$} faster, with \textbf{84\%} MAC and \textbf{83\%} parameter reductions compared to its teacher, TSD-SR, yet preserves strong perceptual quality (\cref{fig:teaser}) and comparable quantitative results (\cref{fig:bubble}) .

Overall, our contribution is summarized as follows:
\begin{itemize}
\item A Real-ISR model called \textbf{TinySR} that achieves \textbf{5.68$\times$} speedup and \textbf{83\%} parameter reduction compared to its teacher, while maintaining a strong perceptual quality.
\item A novel depth pruning method that incorporates \textbf{Dynamic Inter-block Activation} and \textbf{Expansion-Corrosion Strategy} to enable more effective pruning decisions with preserved model performance.
\item Component streamlining strategies incorporate lightweight VAE, redundant conditional structures pruning, and modulation parameters pre-caching.
\end{itemize}
\section{Related Work}
\label{sec:related}
{\bf Real-World Image Super-Resolution.}
Real-world image super-resolution (Real-ISR) addresses the challenges of reconstructing high-resolution images from low-quality inputs affected by complex, unknown degradations. 
Early approaches like BSRGAN \cite{zhang2021designing} and Real-ESRGAN \cite{wang2021real} pioneered synthetic degradation modeling using random blur, noise, and compression patterns to enhance generalization. 
While these methods improved the model's performance, they often introduced undesirable artifacts.
The emergence of diffusion models, particularly Stable Diffusion \cite{rombach2022high, esser2024scaling}, marked a significant advancement in perceptual quality. 
Techniques incorporating StableSR \cite{wang2023exploring} , DiffBIR \cite{lin2023diffbir} and SeeSR \cite{wu2024seesr} demonstrated remarkable results in SR tasks, but their iterative denoising process rendered them impractical for time-sensitive applications.
Recent efforts have focused on distilling multi-step diffusion processes into efficient one-step networks. 
Real-ISR methods such as OSEDiff \cite{wu2024one} and TSD-SR \cite{dong2024tsd} introduced specialized distillation techniques for this purpose. 
Nevertheless, these models still inherit the substantial computational overhead of their diffusion backbones, with parameter counts often exceeding one billion. 
This high complexity poses a significant challenge for their deployment on mobile and edge devices.

\medskip
\noindent
{\bf Efficient Pruning and Compression Techniques.} 
The deployment of large diffusion models on resource-constrained hardware necessitates efficient model compression techniques. 
TinyFusion \cite{fang2024tinyfusion} enables real-time generation by using learnable depth pruning, which is optimized with LoRA-based fine-tuning and Gumbel-Softmax sampling \cite{jang2016categorical}.
Other methods, such as BK-SDM \cite{kim2024bk} and SnapFusion \cite{li2023snapfusion}, reduce model size and latency through structural pruning and on-the-fly architecture modification. 
In the domain of super-resolution, AdcSR \cite{chen2024adversarial} introduces an adversarial compression methodology that achieves a 3.7$\times$ speedup and a 74\% reduction in parameters, while preserving output quality through well-designed adversarial distillation.
Collectively, these methods represent a substantial advancement in model compression, enabling resource-constrained hardware to generate high-quality outputs with improved computational efficiency.
\section{Methodology}
\label{sec:method}
\begin{figure}[!t]
  \centering
    \includegraphics[width=\linewidth]{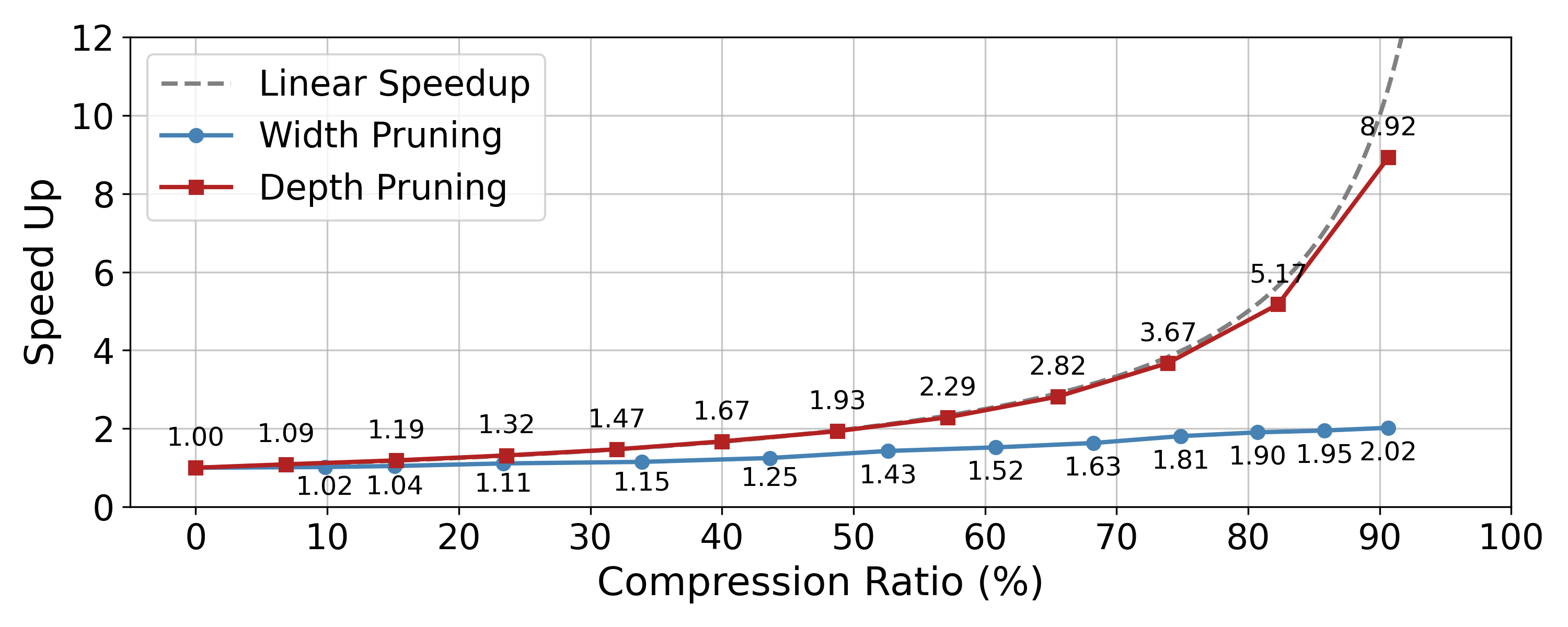}
   \caption{Depth pruning closely aligns with the theoretical linear acceleration curve compared with width pruning.
   }
   \label{fig:speed}
\end{figure}

We aim to learn a compact but effective diffusion model for real-world image super-resolution (Real-ISR). 
To mitigate the computational load and inference latency caused by the excessively deep architecture of the TSD-SR network, we prioritize depth pruning (\cref{subsec:depth}) due to its superior pruning efficiency, as shown in \cref{fig:speed}. 
In \cref{subsec:conpo}, we introduce several key innovations to improve computational efficiency, including a lightweight VAE, a conditional information removal strategy, and a pre-caching technology.
Last, the complete training recipes are given in \cref{subsec:train}.
\begin{figure*}[!ht]
  \centering
    \includegraphics[width=\linewidth]{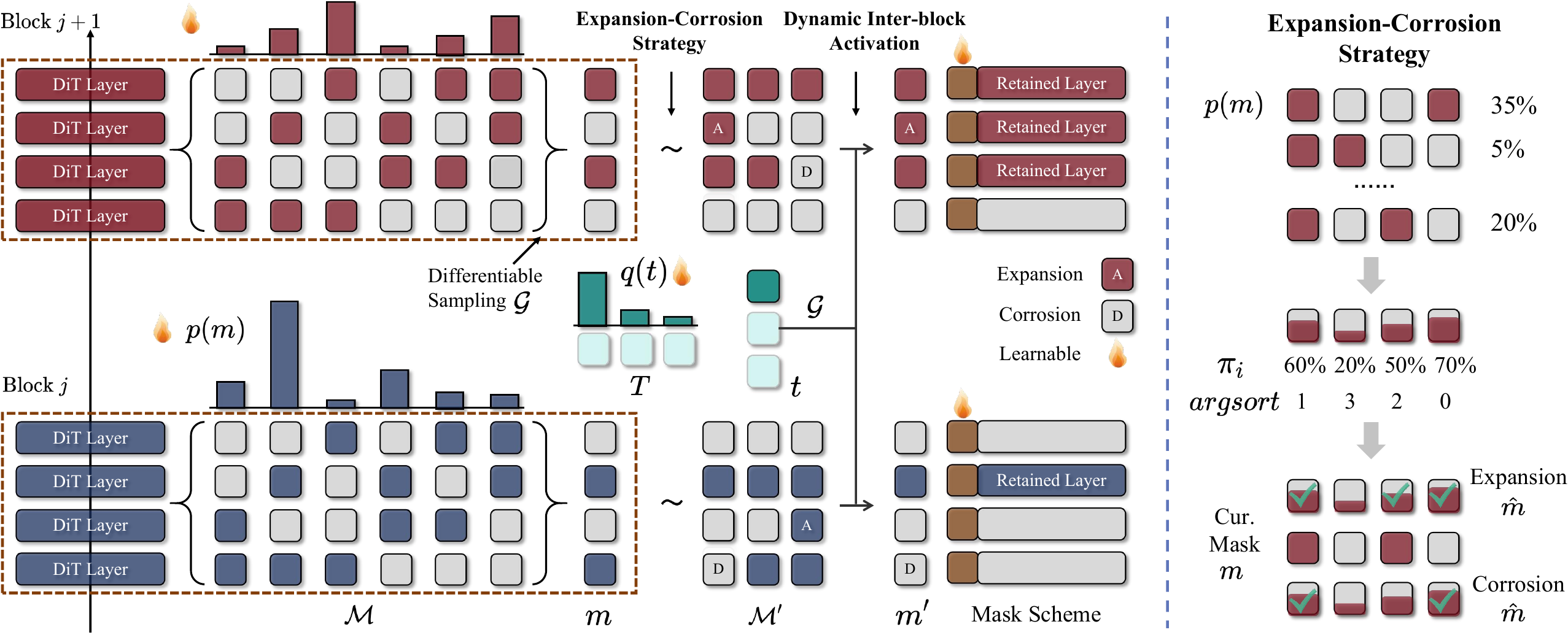}
   \caption{Our proposed mask learning method performs a probability-based decision for candidate solutions, jointly optimized with network weight updates. 
   $p(m)$ characterizes the probability distribution for each block's pruning scheme.
   We perform a transformation probability $q(t)$ to facilitate dynamic interaction between masks in different blocks, namely \textbf{Dynamic Inter-block Activation.} 
   % Building upon the transfer scheme, an \textbf{Expansion-Corrosion Strategy} is employed to determine the final mask based on  maximum marginal probability.
   Leveraging the transfer scheme, we employ an \textbf{Expansion-Corrosion Strategy} to determine the final mask by expanding on elements with maximum marginal probability and corroding those with low marginal probability.
   }
   \label{fig:pipeline}
\end{figure*}
\subsection{Dynamic Depth Pruning}
\label{subsec:depth}
{\bf Depth Pruning Formulation.}
Consider an $ N $-layers transformer parameterized by $ \Phi = [\phi_1, \phi_2, \dots, \phi_N]^T $, where each $ \phi_i \in \mathbb{R}^D $. 
Our goal is to identify an optimal binary mask $ m \in \{0, 1\}^N $ that enables effective pruning while maintaining strong Super-Resolution (SR) performance.
The pruning mechanism is defined by:
\begin{equation}
\label{eq:mask}
x_{i+1} = m_i \cdot \phi_i(x_i) + (1 - m_i) \cdot x_i,
\end{equation}
here, $x_i$ is the input to the $i$-th layer, and $\phi_i(x_i) $ is its output. 
Traditional methods for determining the masking scheme commonly rely on heuristic-based layer importance (e.g., sensitivity analysis and metrics-based selection) or empirical manual configurations. 
However, these carefully designed strategies typically overlook the intricate and interdependent layer dynamics within SR diffusion transformers, leading to suboptimal optimization and potentially weak recoverability in the retained layers.

Inspired by TinyFusion \cite{fang2024tinyfusion}, instead of pursuing models that rely on immediate high-importance feedback, we propose identifying candidate layers with strong recoverability, enabling more efficient teacher-student knowledge distillation.
We formalize the pruning process as a bi-level optimization problem:
\begin{equation}
\label{eq:op_obj2}
\min_{p(m)} \min_{\Delta \Phi} \mathbb{E}_{x, m\sim \mathcal{G}(p(m)) } \left[ \mathcal{L}(x, \Phi + \Delta \Phi,  m) \right].
\end{equation}
This objective is to identify an optimal mask that minimizes the loss function $\mathcal{L}$ during the optimization process.
Since discrete mask selection is non-differentiable, we reparameterize each mask option with a learnable probability parameter $p(m)$ and then use Gumbel-Softmax \cite{jang2016categorical} trick $\mathcal{G}$ to do differentiable sampling:
\begin{equation}
\label{eq:gumbel}
\mathcal{G}(p(m)) = \text{one-hot}\left( \frac{\exp((g_i + \log p_i)/\tau)}{\sum_j \exp((g_j + \log p_j)/\tau)} \right),
\end{equation}
where $g_i$ is random noise drawn from the Gumbel distribution and $\tau$ refers to the temperature term.
The probabilistic sampling of $m$ can be achieved by $m = \mathcal{G}(p(m))^T \cdot \mathcal{M}$, $\mathcal{M}$ represents the complete search space.
For the final pruning decision, we pick $m$ with maximum $p(m)$, as it reveals the strongest recovery. We call this process mask learning.

\medskip
\noindent
{\bf Search Space Dilemma.}
A significant challenge in standard mask learning arises from the combinatorial explosion. 
Let $M:N$ denote the selection of $M$ from $N$ layers, and $C_{N}^{M}$ denote the set of combinations. As $N $ grows, pruning 50\%  will show an explosive trend.
For instance, pruning 50\% of a 24-layer transformer  $C_{24}^{12}$ leads to 2,704,156 possible solutions, making direct probabilistic optimization expensive. 
Naive approaches address this by partitioning an $N$-layer network into $K$ non-overlapping blocks of size $B$, enforcing uniform pruning within each.
Assuming statistical independence of the pruning decisions, the total probability $p(m)$ can be factored into the product of local probabilities $p(m^{(j)})$ :
\begin{equation}
\label{eq:pi_pm}
p(m) = \prod_{j=1}^{K} p(m^{(j)}).
\end{equation}
While this simplifies the search, it drastically curtails the space of possible solutions. 
The cardinality of this feasible subspace, denoted $\mathcal{M}_{valid}$ is given by:
\begin{equation}
\label{eq:valid}
\mathcal{M}_{\text{valid}} = \left\{ m \in \mathcal{M} \middle| \forall j \in \{1,...,K\}, \sum_{i=(j-1)B+1}^{jB} m_i = s \right\}
\end{equation}
The accessible fraction of the search space is $\frac{|\mathcal{M}_{\text{valid}}|}{|\mathcal{M}|} = \frac{46,656}{2,704,156} \approx 1.725\%$. 
This demonstrates that over 98\% of all possible pruning masks are rendered unreachable by the imposition of this seemingly innocuous local constraint.
Consequently, the true optimal pruning scheme may be inadvertently excluded from this narrowly defined search space.

\medskip
\noindent
{\bf Dynamic Inter-block Activation.}
To transcend the limitations of this static partitioning, we introduce a novel pruning framework based on dynamic inter-block activation, as shown in \cref{fig:pipeline}. 
The core motivation is to relax the rigid local constraints and empower the pruning process to discover its own optimal layer distribution across the network. 
Instead of confining the search to the highly-constrained $\mathcal{M}_{valid}$, our method begins within this space yet is allowed to explore beyond it. 
We achieve this by defining a set of probabilistic transformation operators, $T$, governed by distribution parameters $q(t)$. 
Specifically, $T_{j \leftrightarrow h}(m,k)$ transforms $m$ into $m^\prime$ by
pruning $k$ active layers from block $j$ (or $h$) while
restoring $k$ layers in $h$ (or $j$).
This mechanism ensures that the total number of active layers remains constant across the pair of blocks $(j,h)$ while dynamically adjusting their individual sparsity profiles. 
The transformation distribution parameters $q(t)$, are also learnable and applicable to sample a specific scheme through $\mathcal{G}$. 
For example, we assume $k=1, h=j+1$ and define the option space as $\mathcal{M}^\prime = \{m^-, m, m^+\}$, where $m$ is the pruning mask sampled by $p(m)$, and $m^-$ (\textit{resp.} $m^+$) denotes $T_{j \rightarrow j+1}(m,1)$  (\textit{resp.} $T_{j \leftarrow j+1}(m,1)$).
The specific mask can be represented as $m^\prime = \mathcal{G}(q(t))^T \cdot\mathcal{M}^\prime$.
And the total probability can be expressed as:
\begin{equation}
\label{eq:pi_pm_hat}
p(m) = \prod_{j=1}^{K} \underbrace{p(m^{(j)}) \cdot q(t^{(j)}) \cdot r^{(j)}}_{p(m^{\prime(j)})} 
\end{equation}
where $r$ is the probability of reaching $m$ through $t$, and is related to expansion-corrosion strategies.

\medskip
\noindent
{\bf Expansion-Corrosion Strategy.}
Instead of random perturbation-based expansion or corrosion, our approach utilizes the maximum marginal probability:
\begin{equation}
\label{eq:marginal}
\pi_i = p(m_i=1) = \sum_{m:m_i=1} p(m)
\end{equation}
which is derived from training priors.
Specifically, we start by sorting the values of $\pi_i$ in descending order. Then we iteratively select candidate layers, corresponding to these sorted values, to form our candidate mask $\hat m$, which is guaranteed to be $|m \lor \hat{m}|_1 - |m |_1 = k$ for expansion and  $|m |_1 - |m \land \hat{m}|_1 = k$ for erosion.
To ensure backpropagation, we replace the original intersection and union operations with the following simplified computations:
\begin{equation}
\label{eq:union_inter}
m^- = m \odot \hat m, \quad m^+ = clamp(m + \hat m , 0, 1),
\end{equation}
here, \( \odot \) represents the element-wise product. This continuous relaxation maintains the meaning of expansion or corrosion while allowing for gradient flow, which is crucial for training our pruning models.

\medskip
\noindent
{\bf Pruning Decision.} 
Unlike traditional mask learning, where decisions are made directly from $p(m)$, our approach determines the transformations for each block based on maximum $q(t)$ and then selects the final mask based on $\pi$.

\subsection{Component Streamlining}
\label{subsec:conpo}
\begin{figure}[!t]
  \centering
    \includegraphics[width=\linewidth]{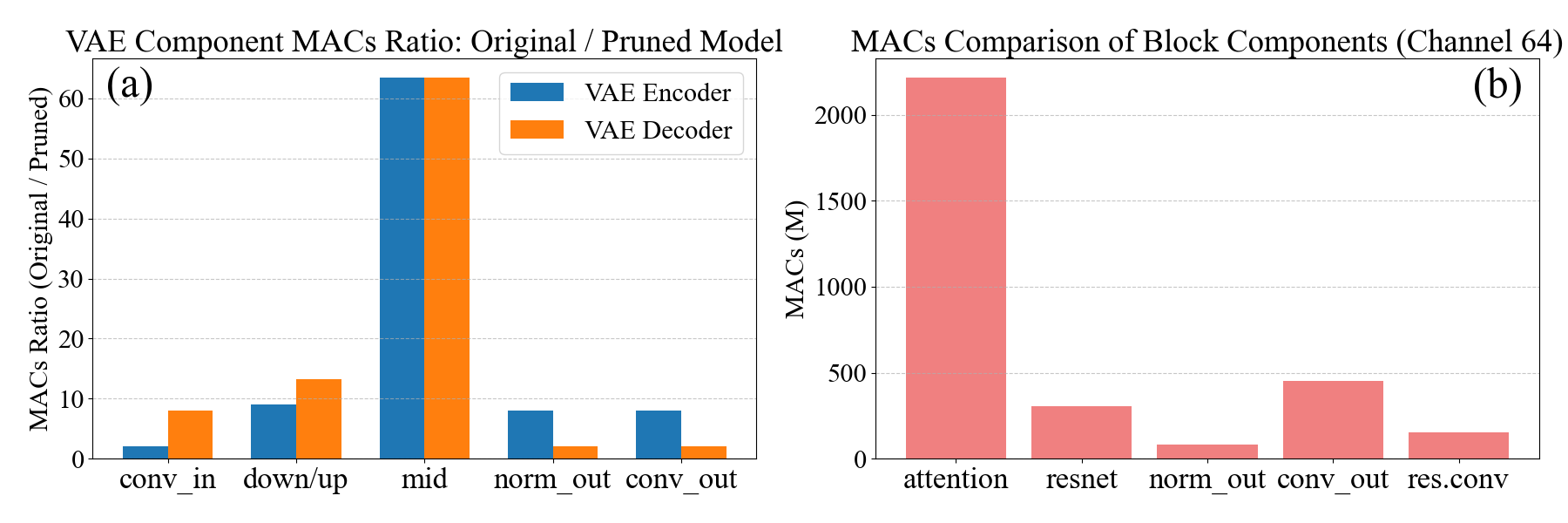}
   \caption{
   Left: Channel pruning effectively reduces MACs across all modules, particularly in the computationally intensive down/up and middle blocks.
   Right: Attention mechanism dominates the computational cost in a 64-channel VAE.
   }
   \label{fig:vae}
\end{figure}

%-------------------------------VAE ----------------------------------------------
{\bf Efficient VAE Architecture.}
We identify three primary performance bottlenecks in the VAE of the TSD-SR framework: 
1) Excessively large channel dimensions. 
As shown in \cref{fig:vae}(a), the MACs of components exhibit significant and widely varying reductions after pruning, particularly across compute-intensive modules such as the down, up and mid blocks.
2) Computationally intensive attention mechanisms. 
As shown in \cref{fig:vae}(b), compared to the resnet block, the attention block poses a more significant computational bottleneck, primarily due to its considerably higher MAC count.
3) Time-consuming standard convolution. 
Convolutional operations are computationally dominant within the VAE, constituting more than 85\% of the overall computational workload.
To address the above problem, we perform pruning on both the encoder and decoder. 
Specifically, following \cite{taesd}, we first prune the maximum channel width to 64 throughout the network, substantially reducing both parameter count and computational complexity.
All self-attention modules are subsequently removed from the architecture to mitigate their substantial computational expense.
To better accommodate higher compression rates, we integrate depthwise separable convolutions \cite{howard2017mobilenets, chollet2017xception} into the encoder. 
However, the same modification proves detrimental to the decoder's performance, leading us to restrict the use of lightweight convolutions to the encoder alone.

\begin{figure*}[!t]
  \centering
    \includegraphics[width=\linewidth]{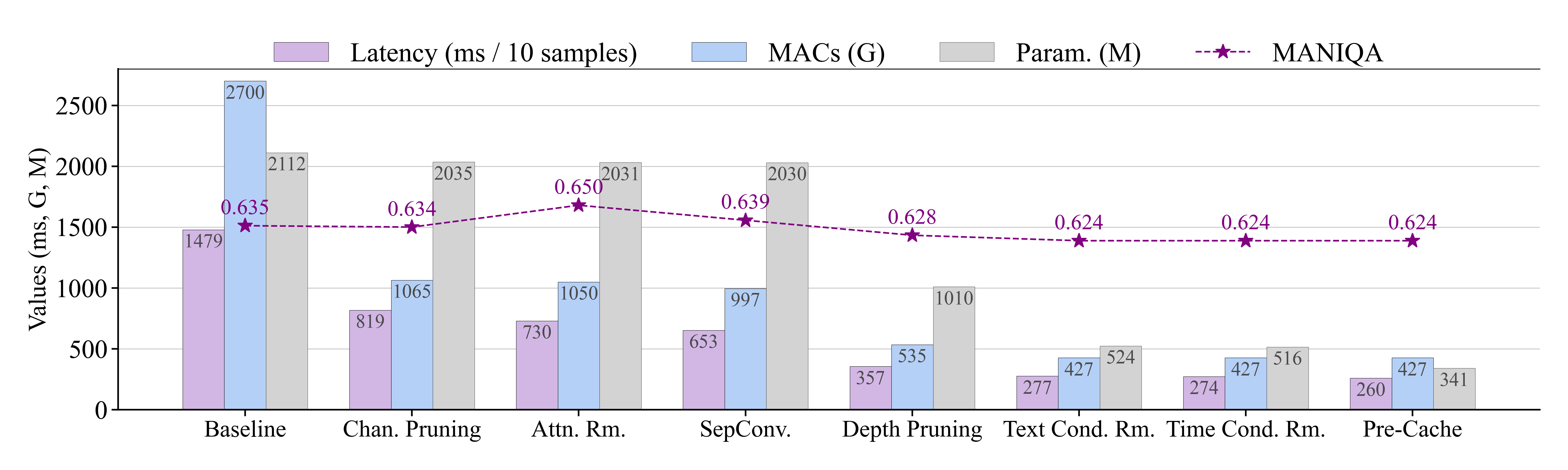}
   \caption{
   Comparisons of performance and efficiency for various design of efficient TinySR. 
   Quality is measured by the MANIQA score, calculated on RealSR dataset. 
   Efficiency is assessed via latency, MACs, and parameters. Latency and MACs are benchmarked for super-resolution a 128 × 128 low-quality image on a n NVIDIA V100 GPU.
    Our model achieves a \textbf{5.68$\times$} acceleration with \textbf{83\%} fewer parameters and \textbf{84\%} lower MACs, while maintaining comparable quality to the baseline.
   }
   \label{fig:latency}
\end{figure*}
\begin{figure}[!t]
    \centering
    % Row 1
    \begin{subfigure}[b]{0.15\textwidth}
        \centering
        \includegraphics[width=\textwidth]{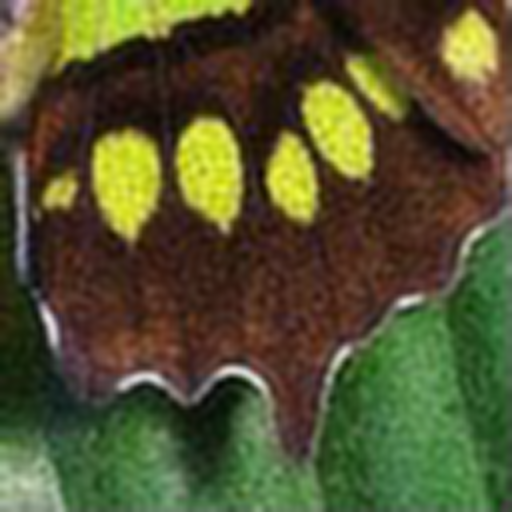} 
        \caption{Low-Quality}
        \label{fig:lr_token}
    \end{subfigure}
    \hfill
    \begin{subfigure}[b]{0.15\textwidth}
        \centering
        \includegraphics[width=\textwidth]{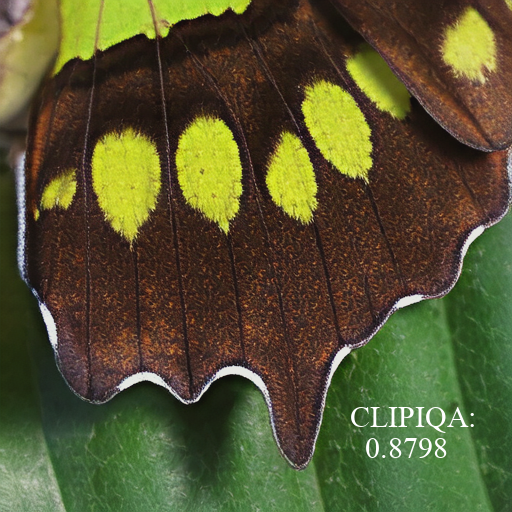} 
        \caption{Baseline}
        \label{fig:tsd_token}
    \end{subfigure}
    \hfill
    \begin{subfigure}[b]{0.15\textwidth}
        \centering
        \includegraphics[width=\textwidth]{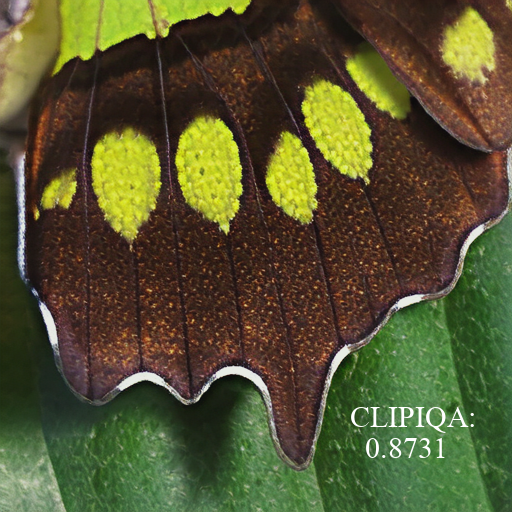} 
        \caption{90\% TP Result}
        \label{fig:out_token}
    \end{subfigure}
    \caption{Applying 90\% token pruning (TP) yields visually comparable results to the baseline with a slight quality drop, indicating the limited contribution of the default prompt.}
    \label{fig:token-pruning}
    \vspace{-0.1em} % Add some vertical space between rows
\end{figure}

%-------------------------------Text Removal ----------------------------------------------
% \subsection{Efficient VAE Architecture}
\medskip
\noindent
{\bf Pruning Redundant Conditional Structures.}
In the TSD-SR model (our teacher), prompt embeddings, essential for text-to-image generation, contribute minimally to image SR tasks (\cref{fig:token-pruning}), as it uses the default prompt embedding. 
Similarly, the time embedding layers, crucial for multi-step diffusion, are redundant in single-step SR and can be removed to reduce computational cost without affecting output quality. 
The removal of redundant structures effectively reduces model inference latency, as shown in \cref{fig:latency}.


%-------------------------------Cache ----------------------------------------------
\medskip
\noindent
{\bf Pre-cache Modulation Param.}
In TSD-SR architecture, shift and scale parameters are generated by adaLN-Zero modulation \cite{esser2024scaling, peebles2023scalable}. 
A key finding is that these modulation parameters generated by our model stabilize post-training and no longer exhibit input dependency. 
This property enables us to pre-compute and cache these parameters, which are then loaded for inference, leading to a significant reduction in computational overhead.

\noindent
Collectively, the above optimizations yield a compact model with an 83\% reduction in parameters and an 84\% reduction in MACs, while maintaining comparable quality and a 5.68$\times$ acceleration, as demonstrated in \cref{fig:latency}.

%-------------------------------Training ----------------------------------------------
\subsection{Training Scheme}
\label{subsec:train}
{\bf VAE Training.}  
We train the VAE encoder $\mathcal{E}_{tiny}$ by aligning latent space features using $\mathrm{MSE}$ loss:
\begin{equation}
\begin{aligned}
\mathcal{L}_{encoder} = \| \mathcal{E}_{tiny}(x_{\mathrm{LR}}) - \mathcal{E}_{pre}(x_{\mathrm{LR}}) \|_2^2,
\end{aligned}
\end{equation}
here, $x_{\mathrm{LR}}$ represents low-quality data, $\mathcal{E}_{\mathrm{pre}}$ is the pre-trained encoder. 
Training is conducted for 100k steps using a batch size of 64 and a learning rate (AdamW optimizer \cite{loshchilov2017decoupled}) of 3e-4 for this phase.
We use $\mathrm{LPIPS}$ loss and $\mathrm{GAN}$ loss to train the VAE decoder $\mathcal{D}_{tiny}$:
\begin{equation}
\begin{aligned}
    \mathcal{L}_{decoder} = 
    \lambda_{\mathrm{1}} \mathcal{L}_{\mathrm{LPIPS}}(\mathcal{D}_{tiny}(\mathcal{E}_{pre}(x_{LR})), x_{HR})  \\
    + \lambda_{\mathrm{2}} \mathcal{L}_{\mathrm{GAN}}(\mathcal{D}_{tiny}(\mathcal{E}_{pre}(x_{LR}))),
\end{aligned}
\end{equation}
here, $x_{\mathrm{HR}}$ represents high-quality data.
We set $\lambda_{\mathrm{1}}$ to 3 and $\lambda_{\mathrm{2}}$ to 1. 
These two loss functions are jointly optimized to ensure both high fidelity and superior perceptual quality in the reconstructed images.
% The Tiny encoder aligns with the teacher's latent representations using $\mathrm{MSE}$ loss, 
% and the Tiny decoder is regularized by $\mathrm{LPIPS}$ and $\mathrm{GAN}$ losses to improve the fidelity and perceptual quality of the reconstructed images. 

\medskip
\noindent
{\bf Pruning Decision Training.}
To ensure the recoverability of pruning, we design the optimization of mask learning using distillation and task losses. 
Specifically, the task loss is defined as $\mathrm{LPIPS}$ loss and $\mathrm{L_1}$ loss is utilized for the distillation loss. 
The total loss is expressed as follows:
\begin{equation}
\begin{aligned}
    \mathcal{L}_{pruning} = &
    \lambda_{\mathrm{3}} \mathcal{L}_{\mathrm{LPIPS}}(\mathcal{D}_{tiny}(z_{stu}), x_{HR})  \\
    & + \lambda_{\mathrm{4}}  \| z_{stu} - z_{tea} \|_1, \\
    where & \quad z_{stu} \sim x_{LR} - \epsilon_{stu}(\mathcal{E}_{tiny}(x_{LR}),t), \\
    & \quad z_{tea} \sim x_{LR} - \epsilon_{tea}(\mathcal{E}_{tea}(x_{LR}),t) ,
    \label{apeq:pruning}
\end{aligned}
\end{equation}
$\epsilon_{stu}$ denotes the student denoising network, while $\epsilon_{tea}$ represents the teacher. $t$ denotes timesteps, and $\mathcal{E}_{tea}$ denotes the teacher encoder.
Training is conducted for 100k iterations across 8 NVIDIA V100 GPUs, employing a learning rate of 5e-5 and a global batch size of 8.
We use LoRA \cite{hu2022lora} for training.
$\lambda_3$ and $\lambda_4$ are both set to 1.

\medskip
\noindent
{\bf Restoration Training.}  
A two-stage training approach is utilized to expedite model convergence. 
The first stage involves latent space training with the distillation loss for efficient feature alignment:
\begin{equation}
\begin{aligned}
    \mathcal{L}_{stage_1} =\| z_{stu} - z_{tea} \|_1 ,
    \label{apeq:stage1}
\end{aligned}
\end{equation}
The meaning of $z_{stu}$ and $z_{tea}$ is the same as mentioned above.
Training in the latent space enables us to use a larger global batch size (128) on 8 V100 GPUs. We set the learning rate to 1e-4 and the LoRA rank to 64. 
The second stage then adds $\mathrm{LPIPS}$ and $\mathrm{GAN}$ losses in the pixel space to improve image-level fidelity:
\begin{equation}
\begin{aligned}
    \mathcal{L}_{stage_2} 
    & = \lambda_{\mathrm{5}}\| z_{stu} - z_{tea} \|_1  \\
    & + \lambda_{\mathrm{6}} \mathcal{L}_{\mathrm{LPIPS}}(\mathcal{D}_{tiny}(z_{stu}), x_{HR})  \\
    & + \lambda_{\mathrm{7}} \mathcal{L}_{\mathrm{GAN}}(\mathcal{D}_{tiny}(z_{stu})),
    \label{apeq:stage2}
\end{aligned}
\end{equation}
where $\lambda_5$, $\lambda_6$, and $\lambda_7$ are set to 5, 1, and 0.3, respectively.
We fine-tune our model for 50k steps on 8 V100 GPUs, with a global batch size of 96, a learning rate of 1e-6 for student (5e-6 for discriminator), and a LoRA rank of 64.

Comprehensive experimental details for the training procedure are available in the supplementary material.

\section{Experiments}
\label{sec:experiment}

\subsection{Experimental Settings}

\begin{table*}[!t]
\caption{Quantitative comparison of various methods on the DIV2K-Val dataset, with all efficiency metrics benchmarked on a NVIDIA V100 GPU. 
The best and second-best results are highlighted in \textbf{bold}, \emph{italic}, respectively.
}
\label{tab:quan}
\resizebox{\linewidth}{!}
{%
\begin{tabular}
{c|ccccccc|cccc}
\toprule
\textbf{Method}  & \textbf{SSIM $\uparrow$} & \textbf{LPIPS $\downarrow$} & \textbf{DISTS $\downarrow$} & \textbf{FID $\downarrow$} & \textbf{NIQE $\downarrow$} & \textbf{MUSIQ $\uparrow$} & \textbf{CLIPIQA $\uparrow$} & \textbf{\#Steps} & \textbf{Time (s)} & \textbf{MACs (G)} & \textbf{\#Param. (M)} \\ \midrule \midrule
StableSR  & 0.5722 & 0.3111 & 0.2046 & \emph{24.95} & 4.7737 & 65.78 & 0.6764 & 200 & 12.731 & 79940 & 1410 \\
DiffBIR   & 0.5717 & 0.3469 & 0.2108 & 33.93 & 4.6056 & 68.54 & 0.7125 & 50 & 6.435 & 24234 & 1717 \\
SeeSR     & 0.6057 & 0.3198 & 0.1953 & 25.81 & 4.8322 & 68.49 & 0.6899 & 50 & 5.722 & 65857 & 2524 \\
ResShift  & \textbf{0.6234 }& 0.3473 & 0.2253 & 42.01 & 6.3615 & 60.63 & 0.5962 & \emph{15} & 0.755 & 5491 & \textbf{174} \\ \midrule
SinSR     & 0.6018 & 0.3262 & 0.2069 & 35.55 & 5.9981 & 62.95 & 0.6501 & \textbf{1} & 0.210 & 2095 & \textbf{174} \\
OSEDiff   & \emph{0.6109} & 0.2942 & 0.1975 & 26.34 & 4.7089 & 67.31 & 0.6681 & \textbf{1} & 0.168 & 2269 & 1761 \\
AdcSR     & 0.6017 & 0.2853 & 0.1899 & 25.52 & 4.3580 & 68.00 & 0.6764 & \textbf{1} & \emph{0.048} & \emph{1049} & 456 \\
TSD-SR    & 0.5808 & \textbf{0.2673} & \textbf{0.1821} & 29.16 & \emph{4.3224} & \textbf{71.69} & \textbf{0.7416} & \textbf{1} & 0.147 & 2700 & 2112 \\
\textbf{Ours} & 0.5725 & \emph{0.2793} & \emph{0.1883} & \textbf{22.44} & \textbf{4.1500} & \emph{69.90} & \emph{0.7201} & \textbf{1} & \textbf{0.026} &  \textbf{427} & \emph{341} \\
\bottomrule
\end{tabular}%
}
\end{table*}

{\bf Datasets.} We utilize DIV2K \cite{agustsson2017ntire}, Flickr2K \cite{timofte2017ntire}, LSDIR \cite{li2023lsdir}, and FFHQ \cite{karras2019style} for training. 
To synthesize low-resolution and high-resolution image pairs, we employ the same degradation pipeline as described in Real-ESRGAN \cite{wang2021real}.
We evaluate the performance of our model on the synthetic DIV2K-Val \cite{agustsson2017ntire} dataset, alongside two real-world datasets, RealSR \cite{cai2019toward} and DRealSR \cite{wei2020component}. 
The datasets consist of paired images with 128x128 low-quality and 512x512 high-quality resolutions.

\medskip
\noindent
{\bf Evaluation Metrics.}
For evaluating our method, we apply both full-reference and no-reference metrics. Full-reference metrics include PSNR and SSIM \cite{wang2004image} (calculated on the Y channel in YCbCr space) for fidelity, LPIPS \cite{zhang2018unreasonable} and DISTS \cite{ding2020image} for perceptual quality, and FID \cite{heusel2017gans} for distribution comparison. No-reference metrics include NIQE \cite{zhang2015feature}, MUSIQ \cite{ke2021musiq}, MANIQA \cite{yang2022maniqa}, CLIPIQA \cite{wang2023exploring}, TOPIQ \cite{chen2024topiq} and Q-Align \cite{wu2023q}.

%----------------------------------Quantitative -------------------------------------------------------------------------
\subsection{Comparison with Existing SR Methods}
We categorize existing outstanding SR models into two groups, single-step and multi-step diffusion models. 
Single-step models include SinSR \cite{wang2024sinsr}, OSEDiff \cite{wu2024one}, TSD-SR \cite{dong2024tsd}, and AdcSR \cite{chen2024adversarial}, 
and multi-step models include StableSR \cite{wang2024exploiting}, DiffBIR \cite{lin2023diffbir}, SeeSR \cite{wu2024seesr}, and ResShift \cite{yue2024resshift}. 
Additional details of GAN-based Real-ISR methods \cite{zhang2021designing,chen2022real,liang2022details,wang2021real} are given in the supplementary material.

\medskip
\noindent
{\bf Quality Comparison.} 
\cref{tab:quan} shows a comparison with DMs-based baselines in Real-ISR tasks. 
For the first four full-reference metrics, our model achieves performance comparable to its teacher TSD-SR and outperforms most other models, securing the second rank for LPIPS and DISTS.
Furthermore, our model achieves the best result for the distribution metric FID. 
For the latter three no-reference metrics, it demonstrates competitive performance: outperforming all other models for NIQE, and ranking second for both MUSIQ and CLIPIQA, thereby surpassing most other methods.

\begin{figure*}[!t]
    \centering

    % Row 1
    \begin{subfigure}[b]{0.16\textwidth}
        \centering
        \includegraphics[width=\textwidth]{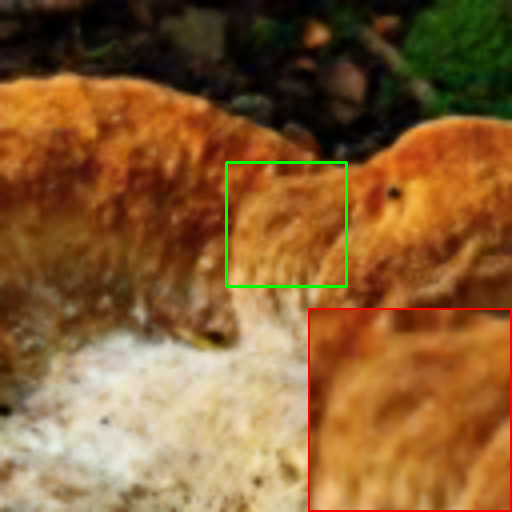} 
        \caption*{Low-Quality}
        \label{fig:row1_vis}
    \end{subfigure}
    \hfill
    \begin{subfigure}[b]{0.16\textwidth}
        \centering
        \includegraphics[width=\textwidth]{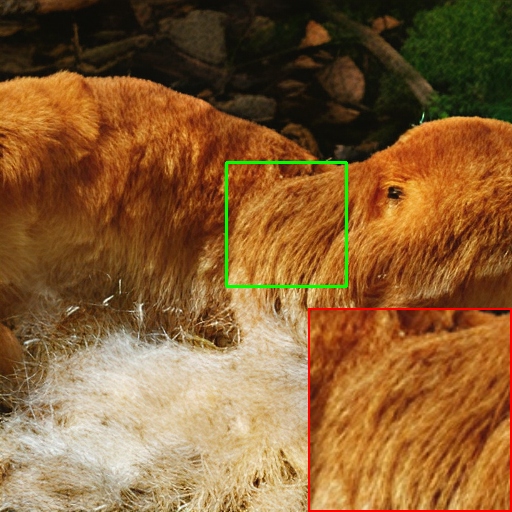} 
        \caption*{StableSR}
        \label{fig:stablesr}
    \end{subfigure}
    \hfill
    \begin{subfigure}[b]{0.16\textwidth}
        \centering
        \includegraphics[width=\textwidth]{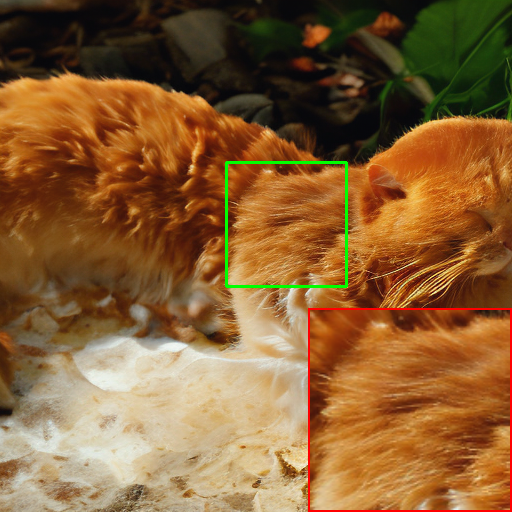} 
        \caption*{SeeSR}
        \label{fig:seesr}
    \end{subfigure}
    \hfill
    \begin{subfigure}[b]{0.16\textwidth}
        \centering
        \includegraphics[width=\textwidth]{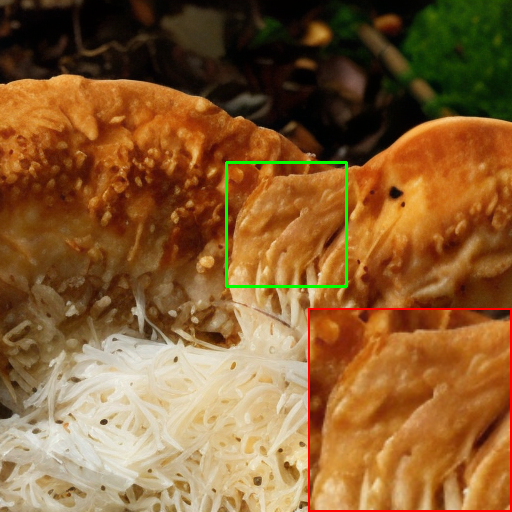} 
        \caption*{DiffBIR}
        \label{fig:diffbir}
    \end{subfigure}
    \hfill
        \begin{subfigure}[b]{0.16\textwidth}
        \centering
        \includegraphics[width=\textwidth]{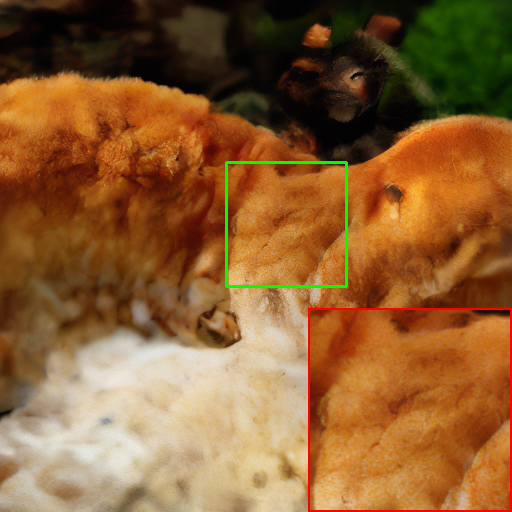} 
        \caption*{ResShift}
        \label{fig:resshift}
    \end{subfigure}
    \hfill
    \begin{subfigure}[b]{0.16\textwidth}
        \centering
        \includegraphics[width=\textwidth]{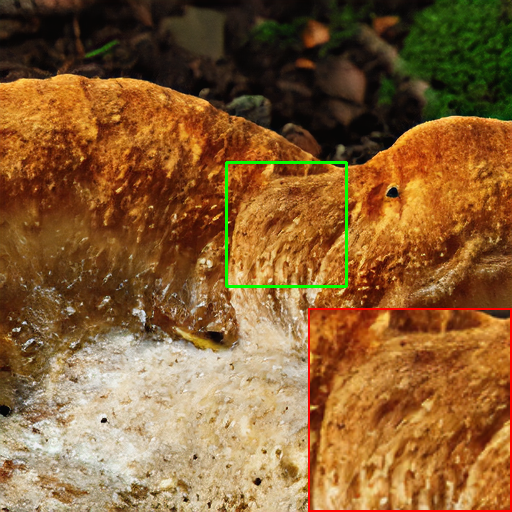} 
        \caption*{Ours}
        \label{fig:ours}
    \end{subfigure}

    % \vspace{1em} % Add some vertical space between rows

    % Row 2
    \begin{subfigure}[b]{0.16\textwidth}
        \centering
        \includegraphics[width=\textwidth]{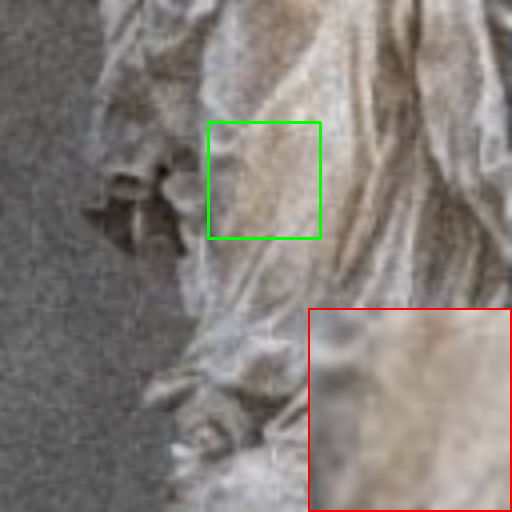} 
        \caption*{Low-Quality}
        \label{fig:lr_row2}
    \end{subfigure}
    \hfill
    \begin{subfigure}[b]{0.16\textwidth}
        \centering
        \includegraphics[width=\textwidth]{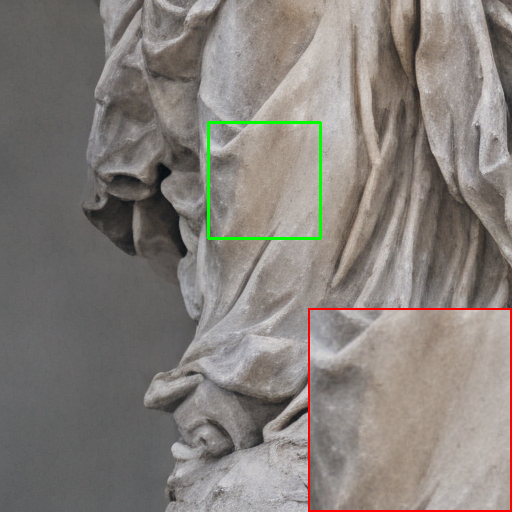} 
        \caption*{OSEDiff}
        \label{fig:osediff}
    \end{subfigure}
    \hfill
    \begin{subfigure}[b]{0.16\textwidth}
        \centering
        \includegraphics[width=\textwidth]{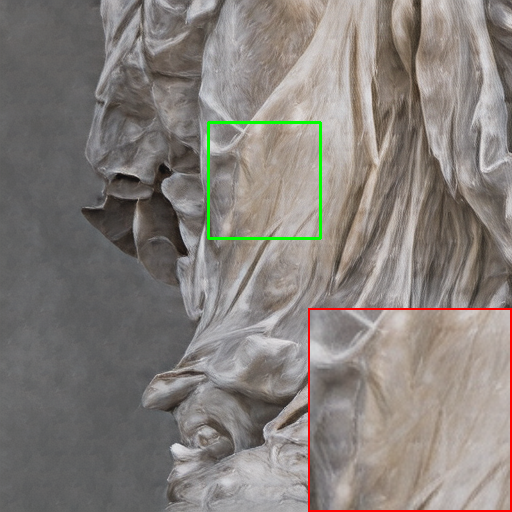} 
        \caption*{TSD-SR}
        \label{fig:tsd-sr}
    \end{subfigure}
    \hfill
    \begin{subfigure}[b]{0.16\textwidth}
        \centering
        \includegraphics[width=\textwidth]{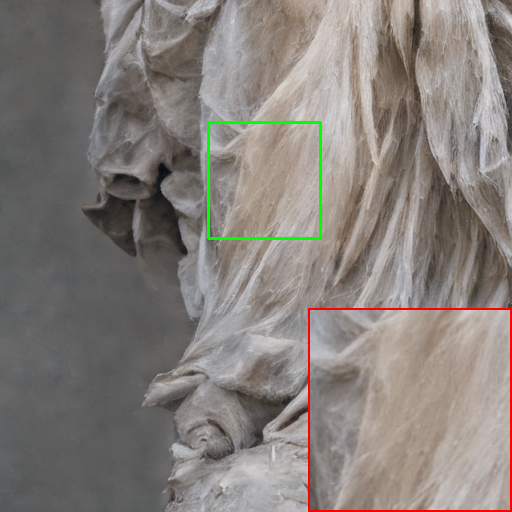} 
        \caption*{AdcSR}
        \label{fig:adcsr}
    \end{subfigure}
    \hfill
    \begin{subfigure}[b]{0.16\textwidth}
        \centering
        \includegraphics[width=\textwidth]{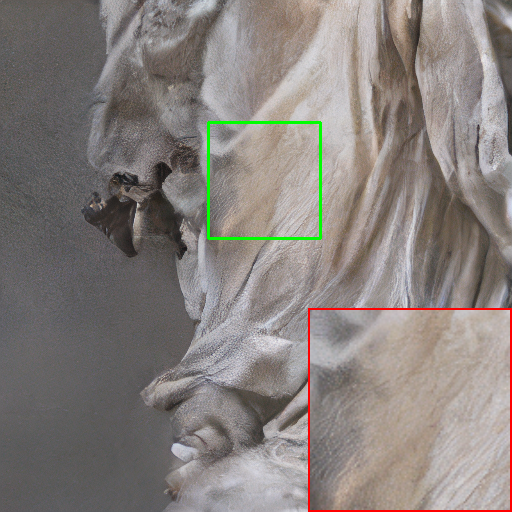} 
        \caption*{SinSR}
        \label{fig:sinsr}
    \end{subfigure}
    \hfill
    \begin{subfigure}[b]{0.16\textwidth}
        \centering
        \includegraphics[width=\textwidth]{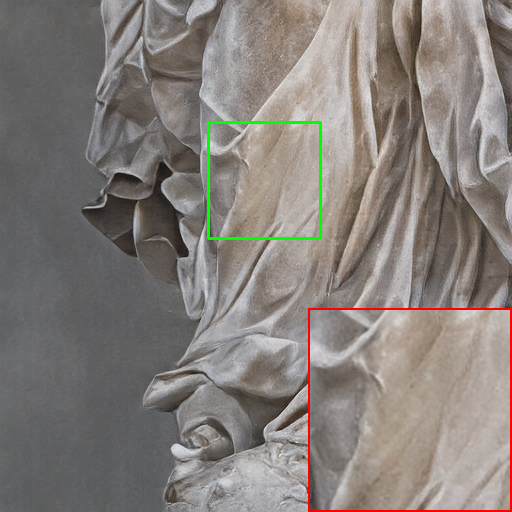} 
        \caption*{Ours}
        \label{fig:ours_vis2}
    \end{subfigure}
    \caption{Qualitative comparisons of different DMs-based Real-ISR methods. 
    Please zoom in for a better view.}
    \label{fig:full_comparison}
\end{figure*}

As illustrated in \cref{fig:teaser} and \cref{fig:full_comparison}, \textbf{TinySR} achieves competitive performance in recovering high-quality, sharp, and photorealistic images.
Visual artifacts, specifically the generation of fake textures, are frequently observed in outputs from multi-step models such as StableSR, SeeSR, DiffBIR, and ResShift due to their propensity for over-generation.
A clear example, demonstrated in \cref{fig:full_comparison} (top), is the spurious generation of hair patterns in regions where they should not exist.
OSEDiff and AdcSR consistently demonstrate suboptimal restoration performance, frequently yielding outputs with discernible blur.
While TSD-SR exhibits excellent generation capabilities, it also tends to produce fake textures, as shown in \cref{fig:full_comparison} (bottom).
\textbf{TinySR}, by contrast, demonstrates robust capabilities in reconstructing natural textures, notably encompassing structural integrity, botanical patterns, and sculpted surface details.

% Speed  -------------------------------------------------------------------------
\medskip
\noindent
{\bf Efficiency Comparison.}
As demonstrated by the last four columns of \cref{tab:quan}, the proposed TinySR exhibits superior efficiency in terms of step number, inference time, and computational cost.
By leveraging one-step inference and advanced compression, our model achieves dramatic efficiency gains over leading multi-step Real-ISR methods while maintaining comparable performance. 
Compared to StableSR, SeeSR, DiffBIR, and ResShift, our model delivers significant speed improvements (489$\times$, 220$\times$, 247$\times$, 29$\times$ ) with corresponding MAC reductions (187$\times$, 154$\times$, 57$\times$, and 12$\times$).
Compared to the one-step model SinSR and OSEDiff, it achieves 8.1$\times$ and 6.4$\times$ acceleration, respectively. 
Compared to its teacher, TSD-SR, it achieves a 5.68$\times$ acceleration, a 84\% reduction in computation, and a 83\% decrease in total parameters, as shown in \cref{fig:latency}.
Notably, In a direct comparison with the current state-of-the-art SR compression model, AdcSR, our model also demonstrates superior efficiency, with a 1.8$\times$ speedup and a 2.45$\times$  computation reduction.

\begin{table}[!t]
\centering
\caption{Performance comparison of depth pruning methods on the DIV2K-Val dataset. Our method exhibits superior recovery performance relative to other pruning strategies.}
\label{tab:ablation_study}
\resizebox{\linewidth}{!}{ 
\begin{tabular}{c|cccccccc}
\toprule
\textbf{Method} & \textbf{DISTS$\downarrow$} & \textbf{FID$\downarrow$} & \textbf{MANIQA$\uparrow$}  & \textbf{CLIPIQA$\uparrow$} & \textbf{TOPIQ$\uparrow$} & \textbf{Q-Align$\uparrow$} \\
\midrule \midrule
Random-Min.       &  0.2012 & 31.40 & 0.5629 & 0.6944 & 0.6314 & 3.5698 \\
Flux-Lite    &  0.2001 & 30.36 & 0.5822 & 0.6983 & 0.6529 & 3.6312  \\
ShortGPT     &  0.2034 & 32.45 & 0.5681 & \emph{0.7116} & \emph{0.6584 }& 3.6110 \\
Sensitivity  &  0.1914 & \emph{26.25} & 0.5879 & 0.6900 & 0.6583 & 3.7135 \\
SnapFusion   &  0.2018 & 31.18 & 0.5788 & 0.7112 & 0.6579 & 3.6468 \\
BK-SDM       &  0.1974 & 29.27 & 0.5792 & 0.6985 & 0.6562 & 3.6466 \\
TinyFusion   &  \emph{0.1904} & 26.46 & \emph{0.5880} & 0.6995 & 0.6580 & \emph{3.7301} \\
\midrule
Ours    & \textbf{0.1883}    & \textbf{22.44}    & \textbf{0.6083}    & \textbf{0.7201}    & \textbf{0.6629}    & \textbf{3.7774}         \\
\bottomrule
\end{tabular}
}
\end{table}
\subsection{Comparison with Depth Pruning Methods}
\label{sec:pruning-strategies}
We evaluate the depth pruning methods following these baselines: 
(1) Perturbation-based – We randomly prune models and select one with minimal task loss (LPIPS) for training; 
(2) Similarity-based – Typically, these methods base their decisions on an analysis of the similarity between each layer's input and output, such as Flux-Lite \cite{flux1-lite} and ShortGPT \cite{men2024shortgpt}; 
(3) Metric-based – Decision-making through metric in general, such as Sensitivity Analysis \cite{han2015learning} and SnapFusion \cite{li2023snapfusion}. 
(4) Experience-based - We follow the design of BK-SDM \cite{kim2024bk} for corresponding pruning; 
(5) Probability-based – Decision by optimized probability parameters, such as TinyFusion \cite{fang2024tinyfusion} and our method.
Randomly generating and then selecting the minimum loss yields a low initial loss, but exhibits extremely weak recovery ability after training.
Similarity-based methods demonstrate suboptimal fidelity (poor DISTS) and pronounced inconsistencies across various no-reference metrics. 
For example, ShortGPT performs well on CLIPIQA but struggles considerably on MANIQA.
Metric-based methods often exhibit a bias towards the metrics they optimize. 
For instance, while Sensitivity Analysis performs well on reference metrics, and SnapFusion excels on CLIPIQA due to its pruning scheme's relation to it, both approaches demonstrate shortcomings in other metrics.
Although BK-SDM and TinyFusion show some effectiveness, our method exhibits enhanced recoverability over all other approaches, performing favorably across both full-reference and no-reference evaluation metrics.

% -----------------------------Ablation Study-------------------------------------------------------------------------
% --------------------------VAE Comparison--------------------------
\subsection{Ablation Study}
{\bf Effect of VAE Compression.}
\cref{tab:ab_vae} presents the ablation studies of VAE compression process.
Channel pruning offers a significant reduction in computational overhead, with only a minor compromise to perceptual reconstruction fidelity.
While removing the attention module effectively doubles inference speed, it can somewhat impact reconstruction quality.
For lightweight convolution, an alternative approach, employing SnapGen's \cite{chen2025snapgen} strategy of expanding channels in \textit{SepConv} intermediate layers, yields performance comparable to our proposed solution. 
However, our method notably exhibit reduced computational overhead and superior inference efficiency.
Our efforts culminated in a lightweight VAE that delivers reconstruction quality on par with the teacher, concurrently achieving a {10$\times$} increase in inference speed and a {22$\times$} reduction in MACs.

\begin{table}[!t]
\caption{Ablation study of VAE compression on DrealSR.  }
\label{tab:ab_vae}
\resizebox{\linewidth}{!}
{
\begin{tabular}{c|ccc|cc}
\toprule
\textbf{Method} & \textbf{PSNR $\uparrow$} & \textbf{SSIM $\uparrow$} & \textbf{LPIPS $\downarrow$}  & \textbf{Time(ms)} & \textbf{MACs(G)} \\ 
\midrule \midrule
Tea. VAE & \textbf{27.77} & \textbf{0.7559} & \textbf{0.2967} & 95.36  & 1781.91 \\
+ Chan. Pruning & 27.49 & \emph{0.7544}  & 0.3052 & 22.79  & 146.56 \\
+ Rm. Attn.  & 27.36 & 0.7420 & 0.3222 & 11.22  & 131.91 \\
SnapGen SepConv & 27.50 & 0.7497 & 0.3054 & \emph{10.75}  & \emph{89.74} \\
\midrule
Ours & \emph{27.56} & 0.7514  &\emph{0.3038} & \textbf{9.25 } & \textbf{78.48} \\
\bottomrule
\end{tabular}
}
\end{table}

% --------------------------Text Comparison--------------------------
\begin{table}[!t]
\caption{Ablation study of removing the text embeddings, time embeddings, and related modules on RealSR.}
\label{tab:ab_cond}
\resizebox{\linewidth}{!}
{
\begin{tabular}{c|cc|ccc}
\toprule
\textbf{Method} & \textbf{CLIPIQA  $\uparrow$} & \textbf{MANIQA  $\uparrow$}  & \textbf{Time (ms)} & \textbf{MACs(G)}   & \textbf{\#Param. (M)} \\ 
\midrule \midrule
w/ Text \& Time  &  \textbf{0.7081} & \textbf{0.6283}  & 35.7 & 535 & 1010 \\
Ours & 0.7035 & 0.6235  & \textbf{27.4} & \textbf{427} & \textbf{516} \\
\bottomrule
\end{tabular}
 }
 % \vspace{-0.1em}
\end{table}
\medskip
\noindent
{\bf Effect of Removing the Text and Time Modules.}
\cref{tab:ab_cond} presents the ablation study on the elimination of text and time conditions, which demonstrates that our approach achieves a highly favorable trade-off between efficiency and quality. 
Excising the text embeddings and related context modules yields a 486M parameter, 108G MACs, and 8ms time reduction, 
while only marginally decreasing the CLIPIQA score by 0.0046 and the MANIQA score by 0.0048.
Subsequent removal of the time modules further reduces parameters by 8M with a negligible impact on final quality. 
% Therefore these modules are omitted from final design.

\section{Conclusion}
\label{sec:conclu}
In this paper, we propose \textbf{TinySR}, a highly efficient model that incorporates several novel contributions. 
For depth pruning, we introduce a \textbf{Dynamic Inter-block Activation} mechanism and an \textbf{Expansion-Corrosion Strategy} 
to facilitate mask learning optimization.
These proposed techniques involve a strategic trade-off between optimization complexity and exploratory potential.
To further reduce computational load, we compress the VAE via channel pruning, attention module removal, and the use of lightweight \textit{SepConv}. 
Finally, we accelerate inference by eliminating time- and prompt-conditioning modules and implementing pre-caching techniques. 
Consequently, \textbf{TinySR} achieves up to a \textbf{5.68$\times$} speedup and a \textbf{83\%} parameter reduction compared to its teacher, TSD-SR, while maintaining high-quality results.
{
    \small
    \bibliographystyle{ieeenat_fullname}
    \bibliography{main}

@String(ICLR = {Int. Conf. Learn. Represent.})

@String(AAAI = {AAAI})

@String(ICLR  = {ICLR})

@article{wu2024one,
  title={One-Step Effective Diffusion Network for Real-World Image Super-Resolution},
  author={Wu, Rongyuan and Sun, Lingchen and Ma, Zhiyuan and Zhang, Lei},
  journal={arXiv preprint arXiv:2406.08177
        
        
        
        
        
        
        
        
},
  year={2024}
}

@article{xie2024addsr,
  title={AddSR: Accelerating Diffusion-based Blind Super-Resolution with Adversarial Diffusion Distillation},
  author={Xie, Rui and Tai, Ying and Zhang, Kai and Zhang, Zhenyu and Zhou, Jun and Yang, Jian},
  journal={arXiv preprint arXiv:2404.01717
        
        
        
        
        
        
        
        
},
  year={2024}
}

@inproceedings{wu2024seesr,
  title={Seesr: Towards semantics-aware real-world image super-resolution},
  author={Wu, Rongyuan and Yang, Tao and Sun, Lingchen and Zhang, Zhengqiang and Li, Shuai and Zhang, Lei},
  booktitle={Proceedings of the IEEE/CVF conference on computer vision and pattern recognition},
  pages={25456--25467},
  year={2024}
}

@inproceedings{wang2021real,
  title={Real-esrgan: Training real-world blind super-resolution with pure synthetic data},
  author={Wang, Xintao and Xie, Liangbin and Dong, Chao and Shan, Ying},
  booktitle={Proceedings of the IEEE/CVF international conference on computer vision},
  pages={1905--1914},
  year={2021}
}

@inproceedings{zhang2021designing,
  title={Designing a practical degradation model for deep blind image super-resolution},
  author={Zhang, Kai and Liang, Jingyun and Van Gool, Luc and Timofte, Radu},
  booktitle={Proceedings of the IEEE/CVF International Conference on Computer Vision},
  pages={4791--4800},
  year={2021}
}

@inproceedings{yu2024scaling,
  title={Scaling up to excellence: Practicing model scaling for photo-realistic image restoration in the wild},
  author={Yu, Fanghua and Gu, Jinjin and Li, Zheyuan and Hu, Jinfan and Kong, Xiangtao and Wang, Xintao and He, Jingwen and Qiao, Yu and Dong, Chao},
  booktitle={Proceedings of the IEEE/CVF Conference on Computer Vision and Pattern Recognition},
  pages={25669--25680},
  year={2024}
}

@inproceedings{wang2024sinsr,
  title={SinSR: diffusion-based image super-resolution in a single step},
  author={Wang, Yufei and Yang, Wenhan and Chen, Xinyuan and Wang, Yaohui and Guo, Lanqing and Chau, Lap-Pui and Liu, Ziwei and Qiao, Yu and Kot, Alex C and Wen, Bihan},
  booktitle={Proceedings of the IEEE/CVF Conference on Computer Vision and Pattern Recognition},
  pages={25796--25805},
  year={2024}
}

@article{yue2024resshift,
  title={Resshift: Efficient diffusion model for image super-resolution by residual shifting},
  author={Yue, Zongsheng and Wang, Jianyi and Loy, Chen Change},
  journal={Advances in Neural Information Processing Systems},
  volume={36},
  year={2024}
}

@article{wang2024exploiting,
  title={Exploiting diffusion prior for real-world image super-resolution},
  author={Wang, Jianyi and Yue, Zongsheng and Zhou, Shangchen and Chan, Kelvin CK and Loy, Chen Change},
  journal={International Journal of Computer Vision},
  pages={1--21},
  year={2024},
  publisher={Springer}
}

@article{lin2023diffbir,
  title={Diffbir: Towards blind image restoration with generative diffusion prior},
  author={Lin, Xinqi and He, Jingwen and Chen, Ziyan and Lyu, Zhaoyang and Dai, Bo and Yu, Fanghua and Ouyang, Wanli and Qiao, Yu and Dong, Chao},
  journal={arXiv preprint arXiv:2308.15070},
  year={2023}
}

@inproceedings{zhang2018unreasonable,
  title={The unreasonable effectiveness of deep features as a perceptual metric},
  author={Zhang, Richard and Isola, Phillip and Efros, Alexei A and Shechtman, Eli and Wang, Oliver},
  booktitle={Proceedings of the IEEE conference on computer vision and pattern recognition},
  pages={586--595},
  year={2018}
}

@inproceedings{liang2022details,
  title={Details or artifacts: A locally discriminative learning approach to realistic image super-resolution},
  author={Liang, Jie and Zeng, Hui and Zhang, Lei},
  booktitle={Proceedings of the IEEE/CVF Conference on Computer Vision and Pattern Recognition},
  pages={5657--5666},
  year={2022}
}

@inproceedings{chen2022real,
  title={Real-world blind super-resolution via feature matching with implicit high-resolution priors},
  author={Chen, Chaofeng and Shi, Xinyu and Qin, Yipeng and Li, Xiaoming and Han, Xiaoguang and Yang, Tao and Guo, Shihui},
  booktitle={Proceedings of the 30th ACM International Conference on Multimedia},
  pages={1329--1338},
  year={2022}
}

@article{wang2004image,
  title={Image quality assessment: from error visibility to structural similarity},
  author={Wang, Zhou and Bovik, Alan C and Sheikh, Hamid R and Simoncelli, Eero P},
  journal={IEEE transactions on image processing},
  volume={13},
  number={4},
  pages={600--612},
  year={2004},
  publisher={IEEE}
}

@article{heusel2017gans,
  title={Gans trained by a two time-scale update rule converge to a local nash equilibrium},
  author={Heusel, Martin and Ramsauer, Hubert and Unterthiner, Thomas and Nessler, Bernhard and Hochreiter, Sepp},
  journal={Advances in neural information processing systems},
  volume={30},
  year={2017}
}

@article{zhang2015feature,
  title={A feature-enriched completely blind image quality evaluator},
  author={Zhang, Lin and Zhang, Lei and Bovik, Alan C},
  journal={IEEE Transactions on Image Processing},
  volume={24},
  number={8},
  pages={2579--2591},
  year={2015},
  publisher={IEEE}
}

@inproceedings{yang2022maniqa,
  title={Maniqa: Multi-dimension attention network for no-reference image quality assessment},
  author={Yang, Sidi and Wu, Tianhe and Shi, Shuwei and Lao, Shanshan and Gong, Yuan and Cao, Mingdeng and Wang, Jiahao and Yang, Yujiu},
  booktitle={Proceedings of the IEEE/CVF Conference on Computer Vision and Pattern Recognition},
  pages={1191--1200},
  year={2022}
}

@inproceedings{ke2021musiq,
  title={Musiq: Multi-scale image quality transformer},
  author={Ke, Junjie and Wang, Qifei and Wang, Yilin and Milanfar, Peyman and Yang, Feng},
  booktitle={Proceedings of the IEEE/CVF international conference on computer vision},
  pages={5148--5157},
  year={2021}
}

@inproceedings{wang2023exploring,
  title={Exploring clip for assessing the look and feel of images},
  author={Wang, Jianyi and Chan, Kelvin CK and Loy, Chen Change},
  booktitle={Proceedings of the AAAI conference on artificial intelligence},
  volume={37},
  number={2},
  pages={2555--2563},
  year={2023}
}

@article{ding2020image,
  title={Image quality assessment: Unifying structure and texture similarity},
  author={Ding, Keyan and Ma, Kede and Wang, Shiqi and Simoncelli, Eero P},
  journal={IEEE transactions on pattern analysis and machine intelligence},
  volume={44},
  number={5},
  pages={2567--2581},
  year={2020},
  publisher={IEEE}
}

@inproceedings{wei2020component,
  title={Component divide-and-conquer for real-world image super-resolution},
  author={Wei, Pengxu and Xie, Ziwei and Lu, Hannan and Zhan, Zongyuan and Ye, Qixiang and Zuo, Wangmeng and Lin, Liang},
  booktitle={Computer Vision--ECCV 2020: 16th European Conference, Glasgow, UK, August 23--28, 2020, Proceedings, Part VIII 16},
  pages={101--117},
  year={2020},
  organization={Springer}
}

@inproceedings{cai2019toward,
  title={Toward real-world single image super-resolution: A new benchmark and a new model},
  author={Cai, Jianrui and Zeng, Hui and Yong, Hongwei and Cao, Zisheng and Zhang, Lei},
  booktitle={Proceedings of the IEEE/CVF international conference on computer vision},
  pages={3086--3095},
  year={2019}
}

@inproceedings{agustsson2017ntire,
  title={Ntire 2017 challenge on single image super-resolution: Dataset and study},
  author={Agustsson, Eirikur and Timofte, Radu},
  booktitle={Proceedings of the IEEE conference on computer vision and pattern recognition workshops},
  pages={126--135},
  year={2017}
}

@inproceedings{li2023lsdir,
  title={Lsdir: A large scale dataset for image restoration},
  author={Li, Yawei and Zhang, Kai and Liang, Jingyun and Cao, Jiezhang and Liu, Ce and Gong, Rui and Zhang, Yulun and Tang, Hao and Liu, Yun and Demandolx, Denis and others},
  booktitle={Proceedings of the IEEE/CVF Conference on Computer Vision and Pattern Recognition},
  pages={1775--1787},
  year={2023}
}

@inproceedings{karras2019style,
  title={A style-based generator architecture for generative adversarial networks},
  author={Karras, Tero and Laine, Samuli and Aila, Timo},
  booktitle={Proceedings of the IEEE/CVF conference on computer vision and pattern recognition},
  pages={4401--4410},
  year={2019}
}

@inproceedings{timofte2017ntire,
  title={Ntire 2017 challenge on single image super-resolution: Methods and results},
  author={Timofte, Radu and Agustsson, Eirikur and Van Gool, Luc and Yang, Ming-Hsuan and Zhang, Lei},
  booktitle={Proceedings of the IEEE conference on computer vision and pattern recognition workshops},
  pages={114--125},
  year={2017}
}

@article{loshchilov2017decoupled,
  title={Decoupled weight decay regularization},
  author={Loshchilov, Ilya and Hutter, Frank},
  journal={arXiv preprint arXiv:1711.05101},
  year={2017}
}

@article{ho2020denoising,
  title={Denoising diffusion probabilistic models},
  author={Ho, Jonathan and Jain, Ajay and Abbeel, Pieter},
  journal={Advances in neural information processing systems},
  volume={33},
  pages={6840--6851},
  year={2020}
}

@inproceedings{rombach2022high,
  title={High-resolution image synthesis with latent diffusion models},
  author={Rombach, Robin and Blattmann, Andreas and Lorenz, Dominik and Esser, Patrick and Ommer, Bj{\"o}rn},
  booktitle={Proceedings of the IEEE/CVF conference on computer vision and pattern recognition},
  pages={10684--10695},
  year={2022}
}

@article{hu2022lora,
  title={Lora: Low-rank adaptation of large language models.},
  author={Hu, Edward J and Shen, Yelong and Wallis, Phillip and Allen-Zhu, Zeyuan and Li, Yuanzhi and Wang, Shean and Wang, Lu and Chen, Weizhu and others},
  journal={ICLR},
  volume={1},
  number={2},
  pages={3},
  year={2022}
}

@inproceedings{esser2024scaling,
  title={Scaling rectified flow transformers for high-resolution image synthesis},
  author={Esser, Patrick and Kulal, Sumith and Blattmann, Andreas and Entezari, Rahim and M{\"u}ller, Jonas and Saini, Harry and Levi, Yam and Lorenz, Dominik and Sauer, Axel and Boesel, Frederic and others},
  booktitle={Forty-first International Conference on Machine Learning},
  year={2024}
}

@misc{kingma2013auto,
  title={Auto-encoding variational bayes},
  author={Kingma, Diederik P and Welling, Max and others},
  year={2013},
  publisher={Banff, Canada}
}

@article{howard2017mobilenets,
  title={Mobilenets: Efficient convolutional neural networks for mobile vision applications},
  author={Howard, Andrew G and Zhu, Menglong and Chen, Bo and Kalenichenko, Dmitry and Wang, Weijun and Weyand, Tobias and Andreetto, Marco and Adam, Hartwig},
  journal={arXiv preprint arXiv:1704.04861
        
        
        
        
        
        
        
        
        
        
        
        },
  year={2017}
}

@article{chen2024adversarial,
  title={Adversarial diffusion compression for real-world image super-resolution},
  author={Chen, Bin and Li, Gehui and Wu, Rongyuan and Zhang, Xindong and Chen, Jie and Zhang, Jian and Zhang, Lei},
  journal={arXiv preprint arXiv:2411.13383
        
        
        
        
        
        
        
        
        
        
        
        
        
        
        
        
        
        
        
        
        
        
        
        
        
        
        
        
        
        
        
        
        
        
        
        },
  year={2024}
}

@article{fang2024tinyfusion,
  title={TinyFusion: Diffusion Transformers Learned Shallow},
  author={Fang, Gongfan and Li, Kunjun and Ma, Xinyin and Wang, Xinchao},
  journal={arXiv preprint arXiv:2412.01199
        
        
        
        
        
        
        
        
        
        
        
        
        
        
        
        
        
        
        
        
        
        
        
        
        
        
        
        
        
        
        
        
        
        },
  year={2024}
}

@inproceedings{dong2024tsd,
  title={Tsd-sr: One-step diffusion with target score distillation for real-world image super-resolution},
  author={Dong, Linwei and Fan, Qingnan and Guo, Yihong and Wang, Zhonghao and Zhang, Qi and Chen, Jinwei and Luo, Yawei and Zou, Changqing},
  booktitle={Proceedings of the Computer Vision and Pattern Recognition Conference},
  pages={23174--23184},
  year={2025}
}

@inproceedings{nichol2021improved,
  title={Improved denoising diffusion probabilistic models},
  author={Nichol, Alexander Quinn and Dhariwal, Prafulla},
  booktitle={International conference on machine learning},
  pages={8162--8171},
  year={2021},
  organization={PMLR}
}

@article{men2024shortgpt,
  title={Shortgpt: Layers in large language models are more redundant than you expect},
  author={Men, Xin and Xu, Mingyu and Zhang, Qingyu and Wang, Bingning and Lin, Hongyu and Lu, Yaojie and Han, Xianpei and Chen, Weipeng},
  journal={arXiv preprint arXiv:2403.03853},
  year={2024}
}

@inproceedings{kim2024bk,
  title={Bk-sdm: A lightweight, fast, and cheap version of stable diffusion},
  author={Kim, Bo-Kyeong and Song, Hyoung-Kyu and Castells, Thibault and Choi, Shinkook},
  booktitle={European Conference on Computer Vision},
  pages={381--399},
  year={2024},
  organization={Springer}
}

@article{li2023snapfusion,
  title={Snapfusion: Text-to-image diffusion model on mobile devices within two seconds},
  author={Li, Yanyu and Wang, Huan and Jin, Qing and Hu, Ju and Chemerys, Pavlo and Fu, Yun and Wang, Yanzhi and Tulyakov, Sergey and Ren, Jian},
  journal={Advances in Neural Information Processing Systems},
  volume={36},
  pages={20662--20678},
  year={2023}
}

@inproceedings{peebles2023scalable,
  title={Scalable diffusion models with transformers},
  author={Peebles, William and Xie, Saining},
  booktitle={Proceedings of the IEEE/CVF international conference on computer vision},
  pages={4195--4205},
  year={2023}
}

@article{jang2016categorical,
  title={Categorical reparameterization with gumbel-softmax},
  author={Jang, Eric and Gu, Shixiang and Poole, Ben},
  journal={arXiv preprint arXiv:1611.01144},
  year={2016}
}

@article{flux1-lite,
  title={Flux.1 Lite: Distilling Flux1.dev for Efficient Text-to-Image Generation},
  author={Daniel Verdú, Javier Martín},
  email={dverdu@freepik.com, javier.martin@freepik.com},
  year={2024},
}

@article{han2015learning,
  title={Learning both weights and connections for efficient neural network},
  author={Han, Song and Pool, Jeff and Tran, John and Dally, William},
  journal={Advances in neural information processing systems},
  volume={28},
  year={2015}
}

@article{dao2022flashattention,
  title={Flashattention: Fast and memory-efficient exact attention with io-awareness},
  author={Dao, Tri and Fu, Dan and Ermon, Stefano and Rudra, Atri and R{\'e}, Christopher},
  journal={Advances in neural information processing systems},
  volume={35},
  pages={16344--16359},
  year={2022}
}

@article{teng2024dim,
  title={Dim: Diffusion mamba for efficient high-resolution image synthesis},
  author={Teng, Yao and Wu, Yue and Shi, Han and Ning, Xuefei and Dai, Guohao and Wang, Yu and Li, Zhenguo and Liu, Xihui},
  journal={arXiv preprint arXiv:2405.14224
        
        },
  year={2024}
}

@article{he2023ptqd,
  title={Ptqd: Accurate post-training quantization for diffusion models},
  author={He, Yefei and Liu, Luping and Liu, Jing and Wu, Weijia and Zhou, Hong and Zhuang, Bohan},
  journal={arXiv preprint arXiv:2305.10657
        
        },
  year={2023}
}

@inproceedings{li2023q,
  title={Q-diffusion: Quantizing diffusion models},
  author={Li, Xiuyu and Liu, Yijiang and Lian, Long and Yang, Huanrui and Dong, Zhen and Kang, Daniel and Zhang, Shanghang and Keutzer, Kurt},
  booktitle={Proceedings of the IEEE/CVF International Conference on Computer Vision},
  pages={17535--17545},
  year={2023}
}

@inproceedings{castells2024ld,
  title={Ld-pruner: Efficient pruning of latent diffusion models using task-agnostic insights},
  author={Castells, Thibault and Song, Hyoung-Kyu and Kim, Bo-Kyeong and Choi, Shinkook},
  booktitle={Proceedings of the IEEE/CVF Conference on Computer Vision and Pattern Recognition},
  pages={821--830},
  year={2024}
}

@article{wang2023patch,
  title={Patch diffusion: Faster and more data-efficient training of diffusion models},
  author={Wang, Zhendong and Jiang, Yifan and Zheng, Huangjie and Wang, Peihao and He, Pengcheng and Wang, Zhangyang and Chen, Weizhu and Zhou, Mingyuan and others},
  journal={Advances in neural information processing systems},
  volume={36},
  pages={72137--72154},
  year={2023}
}

@inproceedings{shi2016real,
  title={Real-time single image and video super-resolution using an efficient sub-pixel convolutional neural network},
  author={Shi, Wenzhe and Caballero, Jose and Husz{\'a}r, Ferenc and Totz, Johannes and Aitken, Andrew P and Bishop, Rob and Rueckert, Daniel and Wang, Zehan},
  booktitle={Proceedings of the IEEE conference on computer vision and pattern recognition},
  pages={1874--1883},
  year={2016}
}

@inproceedings{chen2025snapgen,
  title={Snapgen: Taming high-resolution text-to-image models for mobile devices with efficient architectures and training},
  author={Chen, Jierun and Hu, Dongting and Huang, Xijie and Coskun, Huseyin and Sahni, Arpit and Gupta, Aarush and Goyal, Anujraaj and Lahiri, Dishani and Singh, Rajesh and Idelbayev, Yerlan and others},
  booktitle={Proceedings of the Computer Vision and Pattern Recognition Conference},
  pages={7997--8008},
  year={2025}
}

@article{chen2024topiq,
  title={Topiq: A top-down approach from semantics to distortions for image quality assessment},
  author={Chen, Chaofeng and Mo, Jiadi and Hou, Jingwen and Wu, Haoning and Liao, Liang and Sun, Wenxiu and Yan, Qiong and Lin, Weisi},
  journal={IEEE Transactions on Image Processing},
  volume={33},
  pages={2404--2418},
  year={2024},
  publisher={IEEE}
}

@article{wu2023q,
  title={Q-align: Teaching lmms for visual scoring via discrete text-defined levels},
  author={Wu, Haoning and Zhang, Zicheng and Zhang, Weixia and Chen, Chaofeng and Liao, Liang and Li, Chunyi and Gao, Yixuan and Wang, Annan and Zhang, Erli and Sun, Wenxiu and others},
  journal={arXiv preprint arXiv:2312.17090
        
        
        
        
        
        
        
        
        
        
        
        
        
        
        
        
        
        },
  year={2023}
}

@misc{taesd,
    author={Ollin Boer Bohan},
    title={Tiny AutoEncoder for Stable Diffusion},
    year={2023},
    howpublished={https://github.com/madebyollin/taesd},
}

@inproceedings{chollet2017xception,
  title={Xception: Deep learning with depthwise separable convolutions},
  author={Chollet, Fran{\c{c}}ois},
  booktitle={Proceedings of the IEEE conference on computer vision and pattern recognition},
  pages={1251--1258},
  year={2017}
}
}
\clearpage
\newpage
\appendix
\counterwithin{figure}{section}
\counterwithin{table}{section}
\counterwithin{equation}{section}
\section*{Supplementary Material}

\section{Implementation Details}
\subsection{Data Processing.}
Following \cite{wu2024one, chen2024adversarial,dong2024tsd,wu2024seesr}, the super-resolution process is conducted with a scale factor of 4, upsampling images from 128$\times$128 to 512$\times$512. To create the degraded dataset, Ground Truth (GT) images are first randomly cropped from their original sources. 
Subsequently, these GT images are synthesized into 128x128 degraded data via the well-established Real-ESRGAN \cite{chen2022real} degradation pipeline, involving various corruptions such as noise, blurring, and compression.
This data processing method is widely adopted and well-established. \cite{wang2024exploiting,xie2024addsr,yue2024resshift,wang2024sinsr}.
Moreover, to minimize memory overhead and accelerate training, we pre-encode the low-quality, high-quality, and teacher-generated references into the VAE's latent representations. These latent representations are then cached, enabling their swift retrieval during the training phase.

\subsection{VAE Training.}
For the standard Teacher VAE, we first reduce the number of channels in all intermediate layers to 64 and remove all attention mechanisms, following \cite{taesd}.
Subsequently, all standard convolutional layers are substituted with depthwise separable convolutions (\textit{SepConv}), in line with the methodology proposed by \cite{howard2017mobilenets, chollet2017xception}.

We train the VAE encoder $\mathcal{E}_{tiny}$ by aligning latent space features using $\mathrm{MSE}$ loss. The training objective is defined as:
\begin{equation}
\begin{aligned}
\mathcal{L}_{encoder} = \| \mathcal{E}_{tiny}(x_{\mathrm{LR}}) - \mathcal{E}_{pre}(x_{\mathrm{LR}}) \|_2^2
\end{aligned}
\end{equation}
Here, $x_{\mathrm{LR}}$ represents low-quality data, $\mathcal{E}_{\mathrm{pre}}$ is the pre-trained encoder. 
Training is conducted for 100k steps using a batch size of 64 and a learning rate (AdamW optimizer) of 3e-4 for this phase.

We use $\mathrm{LPIPS}$ loss and $\mathrm{GAN}$ loss to train the VAE decoder $\mathcal{D}_{tiny}$:
\begin{equation}
\begin{aligned}
    \mathcal{L}_{decoder} = 
    \lambda_{\mathrm{1}} \mathcal{L}_{\mathrm{LPIPS}}(\mathcal{D}_{tiny}(\mathcal{E}_{pre}(x_{LR})), x_{HR})  \\
    + \lambda_{\mathrm{2}} \mathcal{L}_{\mathrm{GAN}}(\mathcal{D}_{tiny}(\mathcal{E}_{pre}(x_{LR})))
\end{aligned}
\end{equation}
Here, $x_{\mathrm{HR}}$ represents high-quality data.
We set $\lambda_{\mathrm{1}}$ to 3 and $\lambda_{\mathrm{2}}$ to 1. 
We employ a learning rate of 5e-4 for the decoder and 1e-5 for the discriminator during training. We train the model for 200k iterations using 64 batch size setting. 
The random seed is set to 80 throughout the training.

\subsection{Pruning Decision Training.}
We initialize our model using the pre-trained weights of TSD-SR \cite{dong2024tsd} for pruning training.
We set the pruning rate to 50\% to establish our baseline model. Following the TinyFusion \cite{fang2024tinyfusion} approach, we retained two out of every four layers and employed a dynamic block-wise activation mechanism between adjacent layers.
Our masks are calculated via the Gumbel-Softmax operation \cite{jang2016categorical}.
During network propagation, calculation for a layer is bypassed if its associated mask value is 0.
We optimize the network and probability parameters using SR's task loss and distillation loss aligned with the teacher features.
Specifically, task loss is defined as $\mathrm{LPIPS}$ loss and $\mathrm{L_1}$ loss is utilized for the distillation loss. 
The total loss is expressed as follows:
\begin{equation}
\begin{aligned}
    \mathcal{L}_{pruning} = &
    \lambda_{\mathrm{3}} \mathcal{L}_{\mathrm{LPIPS}}(\mathcal{D}_{tiny}(z_{stu}), x_{HR})  \\
    & + \lambda_{\mathrm{4}}  \| z_{stu} - z_{tea} \|_1 \\
    where & \quad z_{stu} \sim x_{LR} - \epsilon_{stu}(\mathcal{E}_{tiny}(x_{LR}),t), \\
    & \quad z_{tea} \sim x_{LR} - \epsilon_{tea}(\mathcal{E}_{tea}(x_{LR}),t) 
    \label{apeq:pruning}
\end{aligned}
\end{equation}
$\epsilon_{stu}$ denotes the student's denoising network, while $\epsilon_{tea}$ represents the teacher's. $t$ denotes timesteps, and $\mathcal{E}_{tea}$ denotes the teacher encoder.
This encoder differs from the pre-trained version $\mathcal{E}_{pre}$ as it is fine-tuned by TSD-SR \cite{dong2024tsd}.

Training is conducted for 100k iterations across 8 NVIDIA V100 GPUs, employing a learning rate of 5e-5 (AdamW optimizer) and a global batch size of 8.
We use LoRA training, with LoRA rank set to 64.
$\lambda_3$ and $\lambda_4$ are both set to 1.

\begin{table*}[!ht]
% \vspace{-5pt}
\centering
\caption{{Quantitative comparison among different GAN-based and diffusion-based Real-ISR approaches on both synthetic and real-world benchmarks.} ``s'' denotes the required number of sampling steps in the diffusion-based method. The best and second-best results are highlighted in \textbf{bold}, \emph{italic}, respectively}
\vspace{-5pt}
\resizebox{\linewidth}{!}{
\begin{tabular}{l|l|ccccccccc}
\toprule
\textbf{Dataset} & \textbf{Method} 
& \textbf{PSNR} $\uparrow$ 
& \textbf{SSIM} $\uparrow$
& \textbf{LPIPS} $\downarrow$
& \textbf{DISTS} $\downarrow$ 
& \textbf{FID} $\downarrow$ 
& \textbf{NIQE} $\downarrow$ 
& \textbf{MUSIQ} $\uparrow$ 
& \textbf{MANIQA} $\uparrow$ 
& \textbf{CLIPIQA} $\uparrow$ \\
\hline \hline
\multirow{13}{*}{DIV2K-Val} & BSRGAN & \emph{24.58} & {0.6269} & 0.3502 & 0.2280 & 49.55  & 4.75 & 61.68  & 0.5071  & 0.5386   \\
& Real-ESRGAN & 24.02 & \textbf{0.6387} & 0.3150 & 0.2123 & 38.87 & 4.83 & 61.06 & 0.5401 & 0.5251 \\
& LDL & 23.83 & \emph{0.6344} & 0.3256 & 0.2227 & 42.29 & 4.86 & 60.04  & 0.5350 & 0.5180 \\
& FeMASR & 23.06 & 0.5887 & 0.3126 & 0.2057 & 35.87 & 4.74 & 60.83 & 0.5074 & 0.5997 \\
\cline{2-11}
& StableSR-s200 & 23.27 & 0.5722 & 0.3111 & 0.2046 & \emph{24.95} & 4.77 & 65.78 & 0.6164 & 0.6764 \\
& DiffBIR-s50 & 23.13 & 0.5717 & 0.3469 & 0.2108 & 33.93 & {4.61} & 68.54 & \textbf{0.6360} & 0.7125 \\
& SeeSR-s50 & 23.73 & 0.6057 & 0.3198 & {0.1953} & 25.81 & 4.83 & {68.49} & \emph{0.6198} & {0.6899} \\
& ResShift-s15 & \textbf{24.71} & 0.6234 & 0.3473 & 0.2253 & 42.01 & 6.36 & 60.63 & 0.5283 & 0.5962 \\
\cline{2-11}
& SinSR-s1 & {24.41} & 0.6018 & 0.3262 & 0.2069 & 35.55 & 6.00 & 62.95 & 0.5430 & 0.6501 \\
& OSEDiff-s1 & 23.72 & 0.6109 & {0.2942} & 0.1975 & 26.34 & 4.71 & 67.31 & 0.6131 & 0.6681 \\
& AdcSR-s1 & 23.74 & 0.6017 & {0.2853} & {0.1899} & {25.52} & {4.36} & 68.00 & 0.6090 & 0.6764 \\
& TSD-SR-s1 & 23.02 & 0.5808 & \textbf{0.2673} & \textbf{0.1821} & {29.16} & \emph{4.32} & \textbf{71.69} & 0.6192 & \textbf{0.7416} \\
& \textbf{TinySR-s1 (Ours)} & 22.76 & 0.5725 & \emph{0.2793} & \emph{0.1883} & \textbf{24.44} & \textbf{4.15} & \emph{69.90} & 0.6083 & \emph{0.7201} \\
\midrule \midrule
\multirow{13}{*}{DRealSR} & BSRGAN & \textbf{28.70} & {0.8028} & {0.2858} & 0.2143 & 155.61 & 6.54 & 57.15 & 0.4847 & 0.5091 \\
& Real-ESRGAN & 28.61 & \emph{0.8051} & \emph{0.2818} & \textbf{0.2088} & 147.66 & 6.70 & 54.27 & 0.4888 & 0.4512 \\
& LDL & 28.20 & \textbf{0.8124} & \textbf{0.2791} & \emph{0.2127} & 155.51 & 7.14 & 53.94 & 0.4894 & 0.4476 \\
& FeMASR & 26.87 & 0.7569 & 0.3156 & 0.2238 & 157.72 & \emph{5.91} & 53.70 & 0.4413 & 0.5633 \\
\cline{2-11}
& StableSR-s200 & 28.04 & 0.7454 & 0.3279 & 0.2272 & 144.15 & 6.60 & 58.53 & 0.5603 & 0.6250 \\
& DiffBIR-s50 & 25.93 & 0.6525 & 0.4518 & 0.2761 & 177.04 & 6.23 & 65.66 & \textbf{0.6296} & 0.6860 \\
& SeeSR-s50 & 28.14 & 0.7712 & 0.3141 & 0.2297 & 146.95 & 6.46 & {64.74} & \emph{0.6022} & 0.6893 \\
& ResShift-s15 & \emph{28.69} & 0.7874 & 0.3525 & 0.2541 & 176.77 & 7.88 & 52.40 & 0.4756 & 0.5413 \\
\cline{2-11}
& SinSR-s1 & 28.38 & 0.7499 & 0.3669 & 0.2484 & 172.72 & 6.96 & 55.03 & 0.4904 & 0.6412 \\
& OSEDiff-s1 & 27.92 & 0.7836 & 0.2968 & 0.2162 & {135.51} & 6.45 & 64.69 & 0.5898 & {0.6958} \\
& AdcSR-s1 & 28.10 & 0.7726 & 0.3046 & 0.2200 & \textbf{134.05} & 6.45 & \emph{66.26} & 0.5927 & {0.7049} \\
& TSD-SR-s1 & 27.77 & 0.7559 & 0.2967 & {0.2136} & \emph{134.98} & \emph{5.91} & \textbf{66.62} & {0.5874} & \textbf{0.7344} \\
& \textbf{TinySR-s1 (Ours)} & 27.48 & 0.7459 & {0.3116} & {0.2204} & {146.70} & \textbf{5.67} & 65.36 & 0.5804 & \emph{0.7094} \\
\midrule \midrule
\multirow{13}{*}{RealSR} & BSRGAN & {26.38} & \textbf{0.7651} & \textbf{0.2656} & {0.2121} & 141.24 & 5.64 & 63.28 & 0.5425 & 0.5114 \\
& Real-ESRGAN & \textbf{26.65} & \emph{0.7603} & \emph{0.2726} & \textbf{0.2065} & 136.29 & 5.85 & 60.45  & 0.5507 & 0.4518 \\
& LDL & 25.28 & {0.7565} & {0.2750} & {0.2119} & 142.74 & 5.99 & 60.92 & 0.5494 & 0.4559 \\
& FeMASR & 25.07 & 0.7356 & 0.2936 & 0.2285 & 141.01 & 5.77 & 59.05 & 0.4872 & 0.5405 \\
\cline{2-11}
& StableSR-s200 & 24.62 & 0.7041 & 0.3070 & 0.2156 & 128.54 & 5.78 & 65.48 & 0.6223 & 0.6198 \\
& DiffBIR-s50 & 24.24 & 0.6650 & 0.3469 & 0.2300 & 134.56 & 5.49 & 68.35 & \textbf{0.6544} & 0.6961 \\
& SeeSR-s50 & 25.21 & 0.7216 & 0.3003 & 0.2218 & 125.10 & {5.40} & {69.69} & \emph{0.6443} & 0.6671 \\
& ResShift-s15 & \emph{26.39} & 0.7567 & 0.3158 & 0.2432 & 149.59 & 6.87 & 60.22 & 0.5419 & 0.5496 \\
\cline{2-11}
& SinSR-s1 & {26.27} & 0.7351 & 0.3217 & 0.2341 & 137.59 & 6.30 & 60.76 & 0.5418 & 0.6163 \\
& OSEDiff-s1 & 25.15 & 0.7341 & 0.2920 & 0.2128 & {123.48} & 5.65 & {69.10} & 0.6326 & {0.6687} \\
& AdcSR-s1 & 25.47 & 0.7301 & 0.2885 & 0.2129 & {118.41} & {5.35} & \emph{69.90} & 0.6360 & {0.6731} \\
& TSD-SR-s1 & 24.81 & 0.7172 & 0.2743 & \emph{0.2104} & \textbf{114.45} & \emph{5.13} & \textbf{71.19} & {0.6347} & \textbf{0.7160} \\
& \textbf{TinySR-s1 (Ours)} & 24.79 & 0.7171 & {0.2806} & {0.2123} & \emph{118.00} & \textbf{4.74} & 69.78 & 0.6235 & \emph{0.7035} \\
\bottomrule
\end{tabular}}
\label{tab:comp_quantitative}
\end{table*}

\subsection{Restoration Training.}
We perform depth pruning on the TSD-SR according to the pruning mask and discard the condition-related components to initialize our student network.
To achieve rapid convergence, we divided the model's training into two stages.
In the first stage, training is exclusively conducted within the latent space. We employ $\mathrm{L_1}$ loss for teacher-student knowledge distillation to align features. 
The formulation of this distillation loss is consistent with that described in \Cref{apeq:pruning}:
\begin{equation}
\begin{aligned}
    \mathcal{L}_{stage_1} =\| z_{stu} - z_{tea} \|_1 
    \label{apeq:stage1}
\end{aligned}
\end{equation}
The meaning of $z_{stu}$ and $z_{tea}$ is the same as mentioned above.
Training in the latent space enables us to use a larger global batch size (128) on 8 V100 GPUs. We set the learning rate to 1e-4 and the LoRA rank to 64. 
We iterate training 150k steps until convergence.

In stage 2, we further enhance the perceptual quality of the results in image space by fine-tuning the model directly within the image domain.
We additionally incorporate $\mathrm{LPIPS}$ loss and $\mathrm{GAN}$ loss to enhance image restoration.
The total loss is expressed as follows:
\begin{equation}
\begin{aligned}
    \mathcal{L}_{stage_2} 
    & = \lambda_{\mathrm{5}}\| z_{stu} - z_{tea} \|_1  \\
    & + \lambda_{\mathrm{6}} \mathcal{L}_{\mathrm{LPIPS}}(\mathcal{D}_{tiny}(z_{stu}), x_{HR})  \\
    & + \lambda_{\mathrm{7}} \mathcal{L}_{\mathrm{GAN}}(\mathcal{D}_{tiny}(z_{stu}))
    \label{apeq:stage2}
\end{aligned}
\end{equation}
The meaning of $z_{stu}$, $z_{tea}$ and $\mathcal{D}_{tiny}$ is the same as mentioned above.
$\lambda_5$, $\lambda_6$, and $\lambda_7$ are set to 5, 1, and 0.3, respectively.
We fine-tune our model for 50k steps on 8 V100 GPUs, with a global batch size of 96, a learning rate of 1e-6 for student (5e-6 for discriminator), and a LoRA rank of 64.
The random seed for the entire training process is set to 80.
And all training is done on \textit{fp16} precision.

\section{More Comparisons on Benchmarks}
\subsection{More Quantitative Comparisons}
We compared GAN-based and diffusion-based methods across various datasets (DIV2K-Val \cite{agustsson2017ntire}, DrealSR \cite{wei2020component}, RealSR \cite{cai2019toward}), with the results presented in \Cref{tab:comp_quantitative}.
We observe that traditional GAN-based approaches \cite{zhang2021designing,chen2022real,liang2022details,wang2021real} generally excel on full-reference metrics, particularly PSNR and SSIM.
However, some studies indicate that PSNR and SSIM often do not accurately reflect fidelity under more complex degradation conditions \cite{xie2024addsr,dong2024tsd,yu2024scaling}.
In most perceptual quality metrics, such as NIQE \cite{zhang2015feature}, MUSIQ \cite{ke2021musiq}, MANIQA \cite{yang2022maniqa} and CLIPIQA \cite{wang2023exploring}, diffusion-based methods demonstrate superior performance compared to these GANs, highlighting their enhanced capability in generating natural textures.
TinySR achieved competitive performance across most metrics, demonstrating comparable results to its teacher model, TSD-SR, and showcasing the robust recoverability of the pruning methods.

\begin{figure}[!t]
    \centering
    % Row 1
    \begin{subfigure}[b]{0.15\textwidth}
        \centering
        \includegraphics[width=\textwidth]{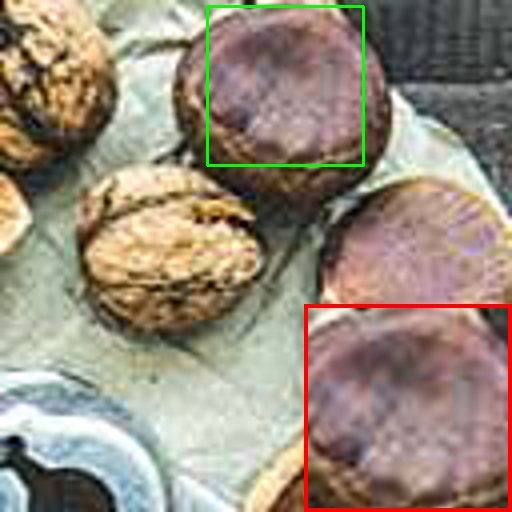} 
    \end{subfigure}
    \hfill
    \begin{subfigure}[b]{0.15\textwidth}
        \centering
        \includegraphics[width=\textwidth]{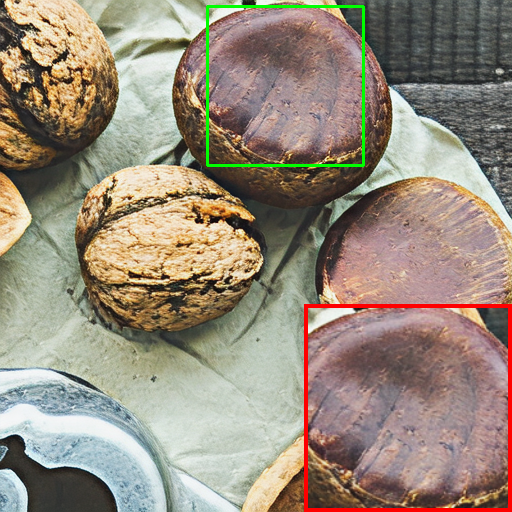} 
    \end{subfigure}
    \hfill
    \begin{subfigure}[b]{0.15\textwidth}
        \centering
        \includegraphics[width=\textwidth]{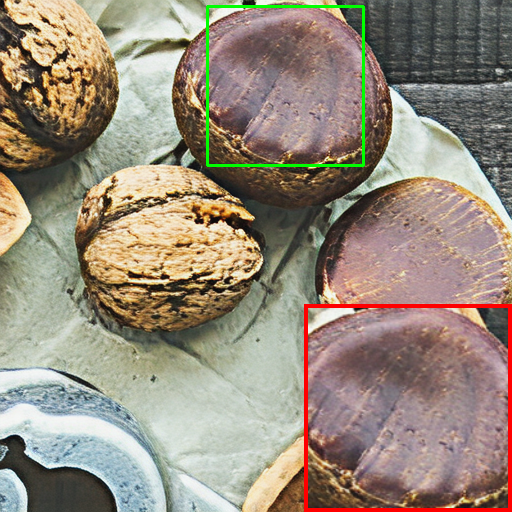} 
    \end{subfigure}

    \vspace{0.5em}
    % Row 2
    \begin{subfigure}[b]{0.15\textwidth}
        \centering
        \includegraphics[width=\textwidth]{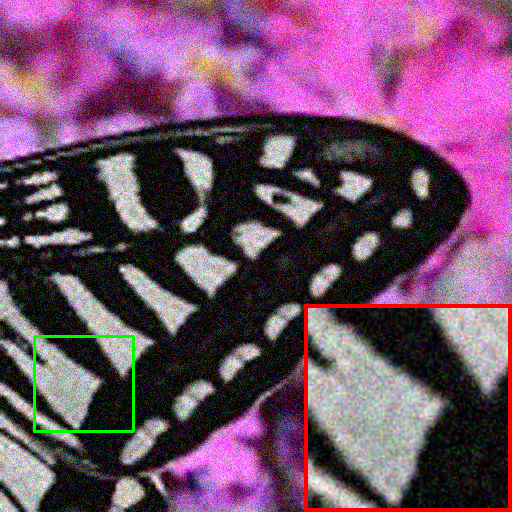} 
        \caption{Low-Quality}
    \end{subfigure}
    \hfill
    \begin{subfigure}[b]{0.15\textwidth}
        \centering
        \includegraphics[width=\textwidth]{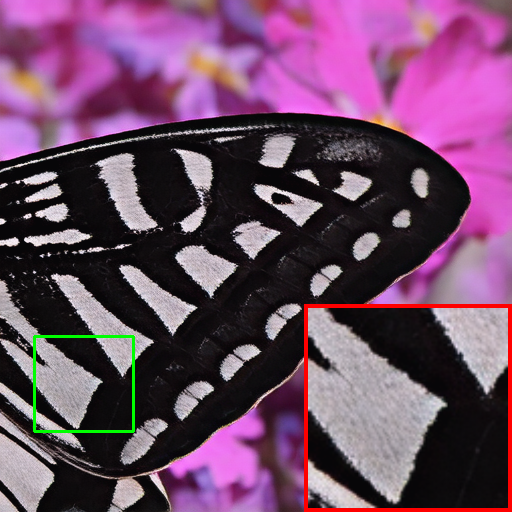} 
        \caption{Baseline}
    \end{subfigure}
    \hfill
    \begin{subfigure}[b]{0.15\textwidth}
        \centering
        \includegraphics[width=\textwidth]{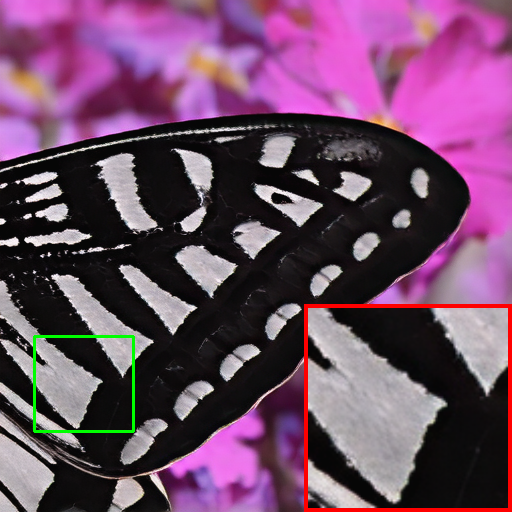} 
        \caption{90\% TP Result}
    \end{subfigure}
    \caption{Applying 90\% token pruning (TP) yields visually comparable results to the baseline with a slight quality drop, indicating the limited contribution of the default prompt.}
    \label{fig:aptoken-pruning}
\end{figure}

\begin{table}[!t]
\centering
\caption{Ablation study of prompt token pruning (TP) on DrealSR dataset. The best is highlighted in \textbf{bold}.}
\label{tab:aptoken}
\resizebox{\linewidth}{!}{ 
\begin{tabular}{l|cccccc}
\toprule
\textbf{Method} & 
\multicolumn{1}{c}{\textbf{PSNR} $\uparrow$} &
\multicolumn{1}{c}{\textbf{LPIPS} $\downarrow$} & 
\multicolumn{1}{c}{\textbf{DISTS} $\downarrow$} & 
\multicolumn{1}{c}{\textbf{NIQE} $\downarrow$} & 
\multicolumn{1}{c}{\textbf{MUSIQ} $\uparrow$} & 
\multicolumn{1}{c}{\textbf{MANIQA} $\uparrow$} \\
\midrule
Baseline
& \textbf{27.77}    & 0.2967    &  0.2136   & \textbf{5.9131}   &  \textbf{66.62} & \textbf{0.5927}     \\
TP 10\% token  
& 27.66 & 0.2945 & 0.2135 & 5.9536 & 66.57 & 0.5870 \\
TP 25\% token  
& 27.65 & 0.2947 & 0.2135 & 5.9510  & 66.53 & 0.5861 \\
TP 50\% token  
& 27.66 & 0.2924 & \textbf{0.2120}  & 5.9350 & 66.35 & 0.5801 \\
TP 75\% token  
& 27.62 & \textbf{0.2860} & 0.2122 & 6.0142  & 65.79 & 0.5783 \\
TP 90\% token  
& 27.59 & 0.2866 & 0.2144 & 6.1461 & 65.01 & 0.5721 \\
\bottomrule
\end{tabular}
}
\end{table}
\begin{figure*}[!t]
    \centering
    % Row 1
    \begin{subfigure}[b]{0.16\textwidth}
        \centering
        \includegraphics[width=\textwidth]{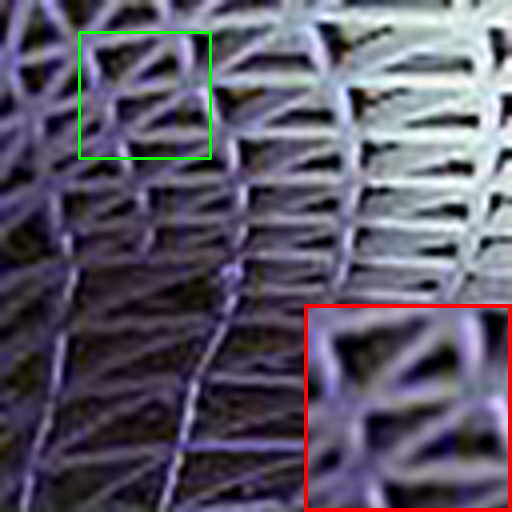} 
        \caption*{Low-Quality}
    \end{subfigure}
    \hfill
    \begin{subfigure}[b]{0.16\textwidth}
        \centering
        \includegraphics[width=\textwidth]{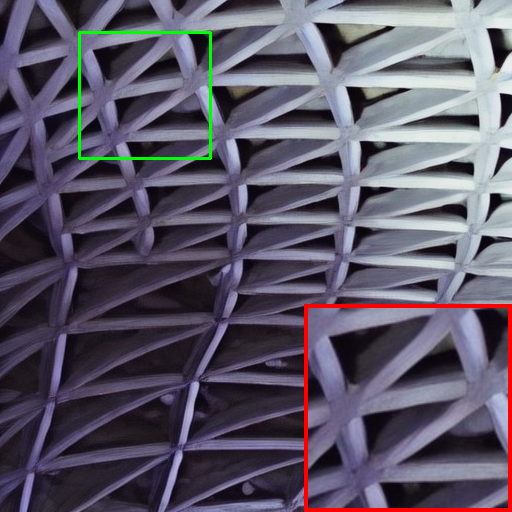}
        \caption*{OSEDiff}
    \end{subfigure}
    \hfill
    \begin{subfigure}[b]{0.16\textwidth}
        \centering
        \includegraphics[width=\textwidth]{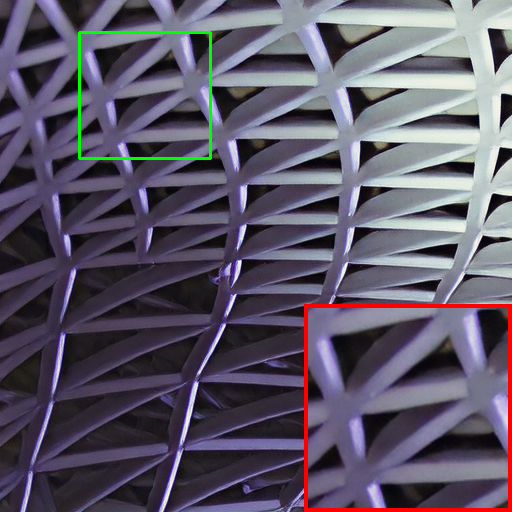} 
        \caption*{TSD-SR}
    \end{subfigure}
    \hfill
    \begin{subfigure}[b]{0.16\textwidth}
        \centering
        \includegraphics[width=\textwidth]{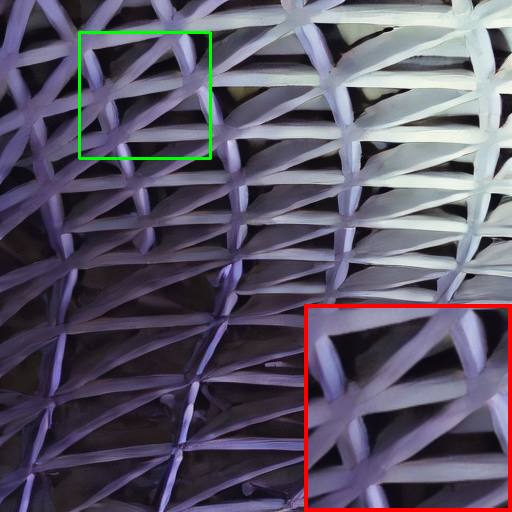} 
        \caption*{AdcSR}
    \end{subfigure}
    \hfill
    \begin{subfigure}[b]{0.16\textwidth}
        \centering
        \includegraphics[width=\textwidth]{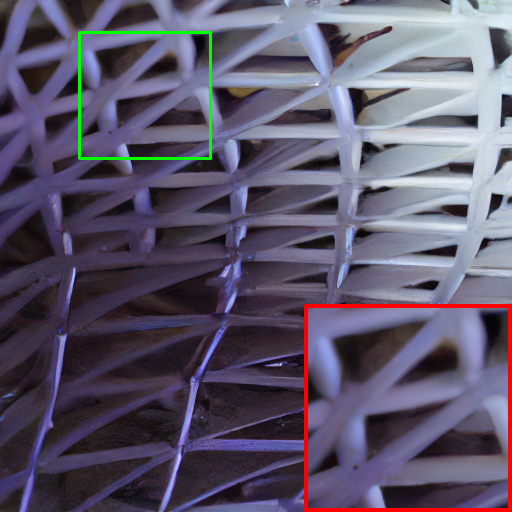} 
        \caption*{SinSR}
    \end{subfigure}
    \hfill
    \begin{subfigure}[b]{0.16\textwidth}
        \centering
        \includegraphics[width=\textwidth]{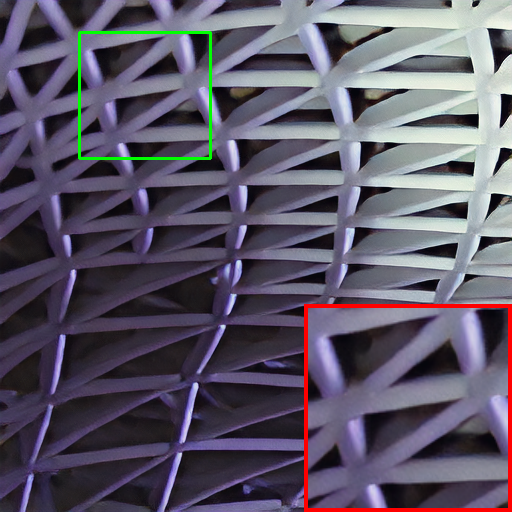} 
        \caption*{Ours}
    \end{subfigure}

    \vspace{0.5em} % Add some vertical space between rows
    % Row2
    \begin{subfigure}[b]{0.16\textwidth}
        \centering
        \includegraphics[width=\textwidth]{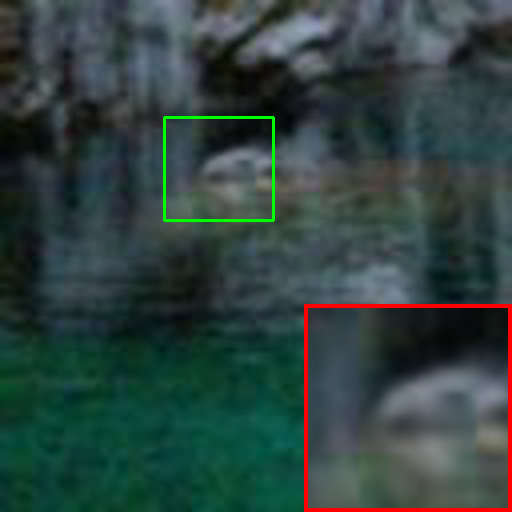} 
        \caption*{Low-Quality}
    \end{subfigure}
    \hfill
    \begin{subfigure}[b]{0.16\textwidth}
        \centering
        \includegraphics[width=\textwidth]{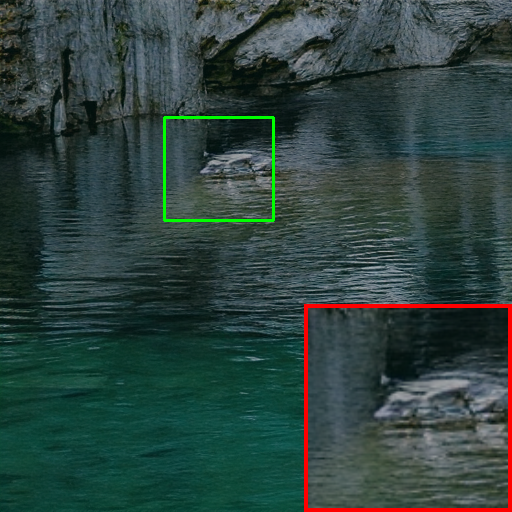} 
        \caption*{StableSR}
    \end{subfigure}
    \hfill
    \begin{subfigure}[b]{0.16\textwidth}
        \centering
        \includegraphics[width=\textwidth]{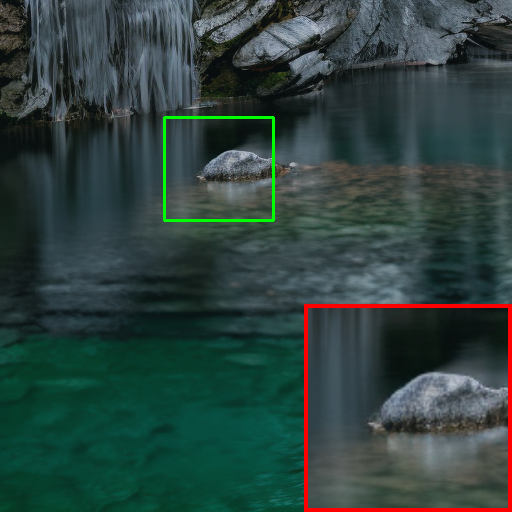} 
        \caption*{SeeSR}
    \end{subfigure}
    \hfill
    \begin{subfigure}[b]{0.16\textwidth}
        \centering
        \includegraphics[width=\textwidth]{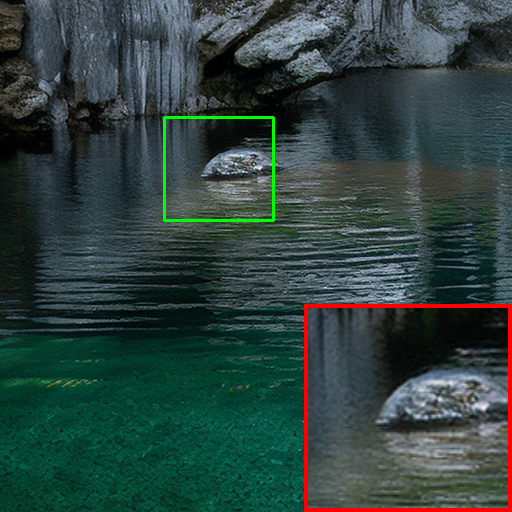} 
        \caption*{DiffBIR}
    \end{subfigure}
    \hfill
    \begin{subfigure}[b]{0.16\textwidth}
        \centering
        \includegraphics[width=\textwidth]{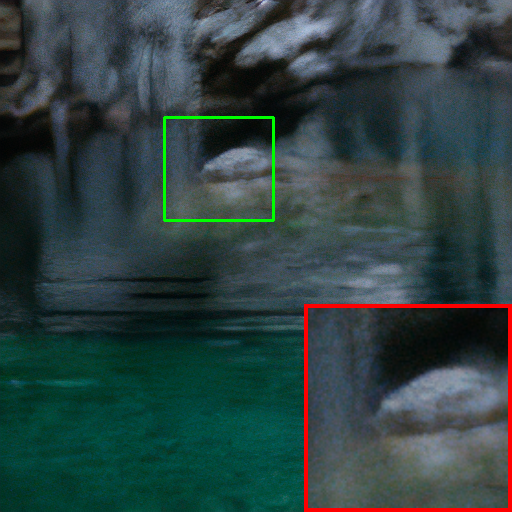} 
        \caption*{ResShift}
    \end{subfigure}
    \hfill
    \begin{subfigure}[b]{0.16\textwidth}
        \centering
        \includegraphics[width=\textwidth]{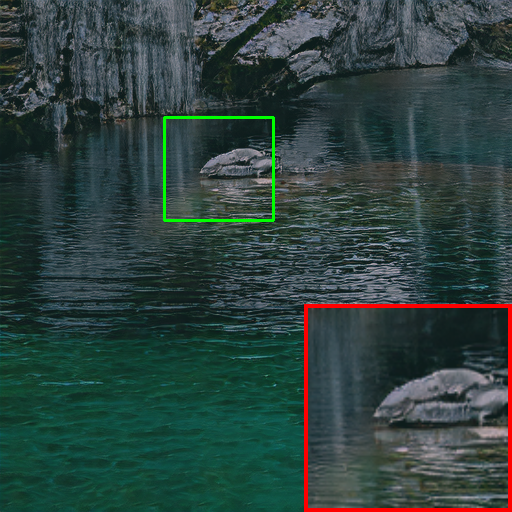} 
        \caption*{Ours}
    \end{subfigure}
    \vspace{0.5em} % Add some vertical space between rows

    % Row 3
    \begin{subfigure}[b]{0.16\textwidth}
        \centering
        \includegraphics[width=\textwidth]{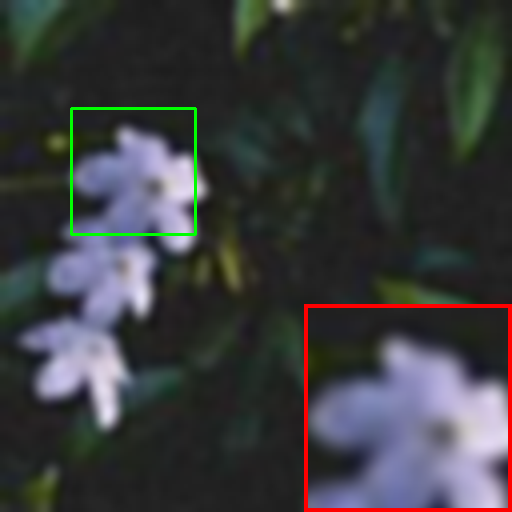} 
        \caption*{Low-Quality}
    \end{subfigure}
    \hfill
    \begin{subfigure}[b]{0.16\textwidth}
        \centering
        \includegraphics[width=\textwidth]{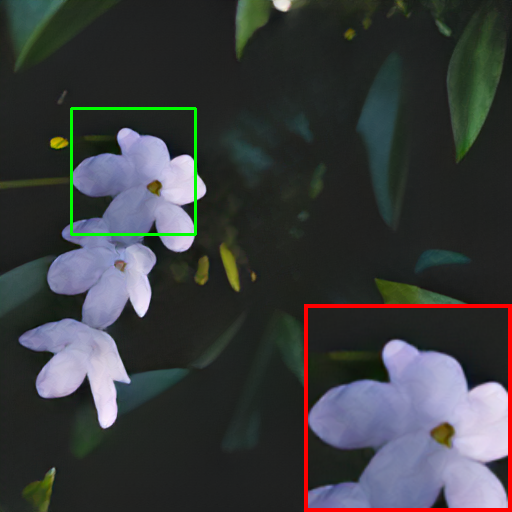} 
        \caption*{BSRGAN}
    \end{subfigure}
    \hfill
    \begin{subfigure}[b]{0.16\textwidth}
        \centering
        \includegraphics[width=\textwidth]{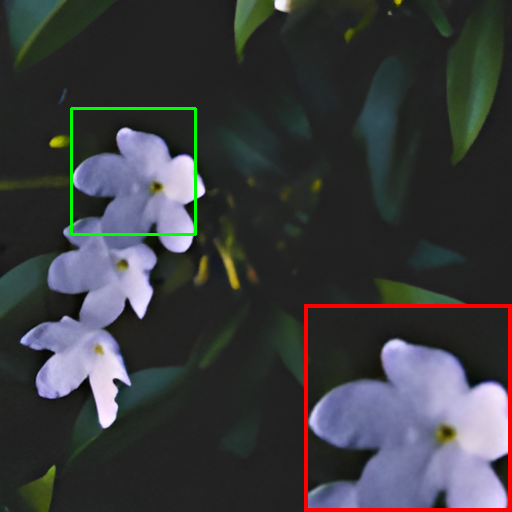} 
        \caption*{Real-ESRGAN}
    \end{subfigure}
    \hfill
    \begin{subfigure}[b]{0.16\textwidth}
        \centering
        \includegraphics[width=\textwidth]{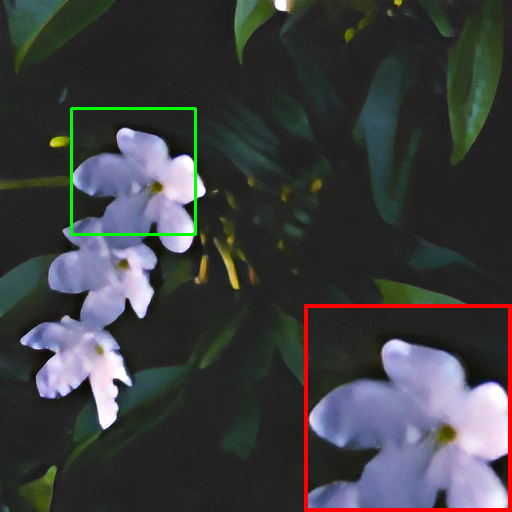} 
        \caption*{FeMASR}
    \end{subfigure}
    \hfill
    \begin{subfigure}[b]{0.16\textwidth}
        \centering
        \includegraphics[width=\textwidth]{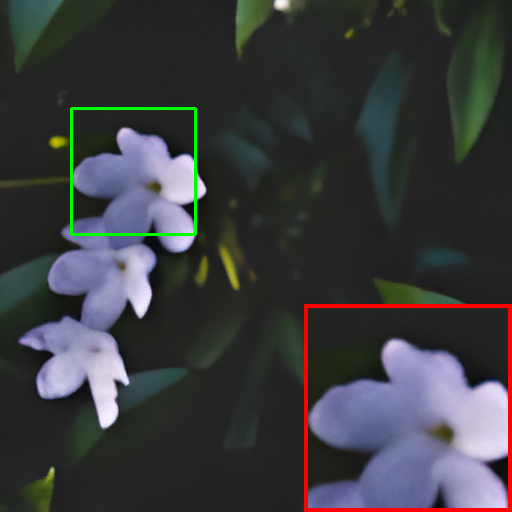} 
        \caption*{LDL}
    \end{subfigure}
    \hfill
    \begin{subfigure}[b]{0.16\textwidth}
        \centering
        \includegraphics[width=\textwidth]{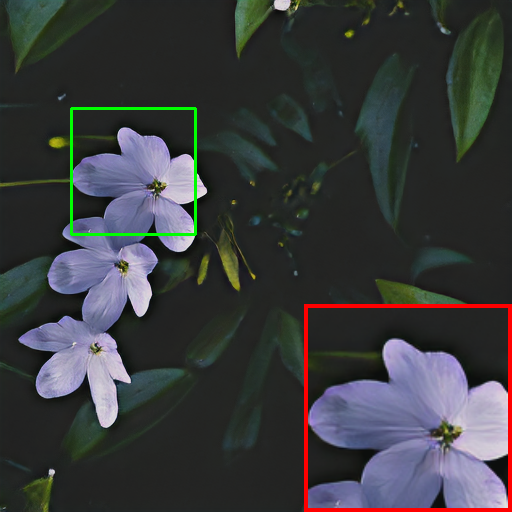} 
        \caption*{Ours}
    \end{subfigure}
    \vspace{0.5em} % Add some vertical space between rows

    % Row 4
    \begin{subfigure}[b]{0.16\textwidth}
        \centering
        \includegraphics[width=\textwidth]{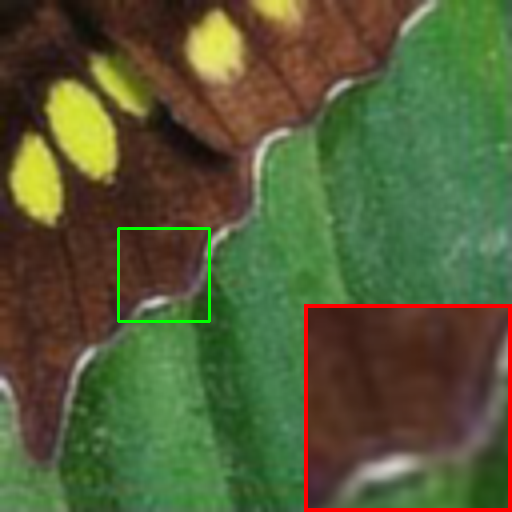} 
        \caption*{Low-Quality}
    \end{subfigure}
    \hfill
    \begin{subfigure}[b]{0.16\textwidth}
        \centering
        \includegraphics[width=\textwidth]{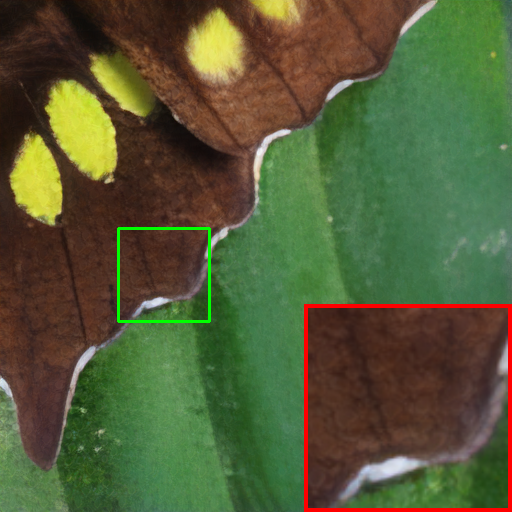} 
        \caption*{BSRGAN}
    \end{subfigure}
    \hfill
    \begin{subfigure}[b]{0.16\textwidth}
        \centering
        \includegraphics[width=\textwidth]{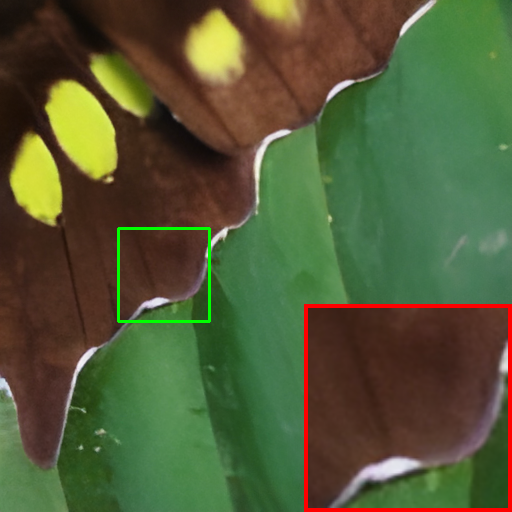} 
        \caption*{LDL}
    \end{subfigure}
    \hfill
    \begin{subfigure}[b]{0.16\textwidth}
        \centering
        \includegraphics[width=\textwidth]{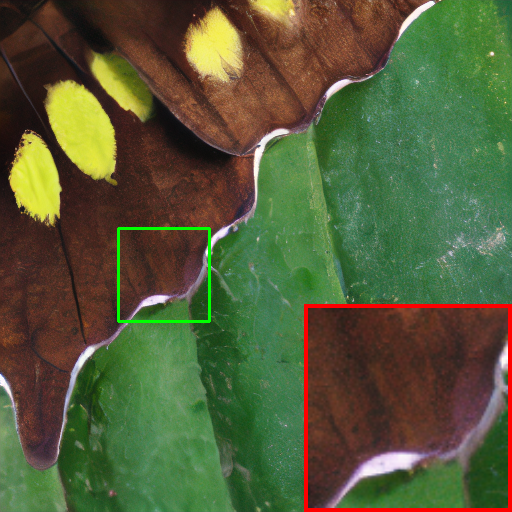} 
        \caption*{ResShift}
    \end{subfigure}
    \hfill
    \begin{subfigure}[b]{0.16\textwidth}
        \centering
        \includegraphics[width=\textwidth]{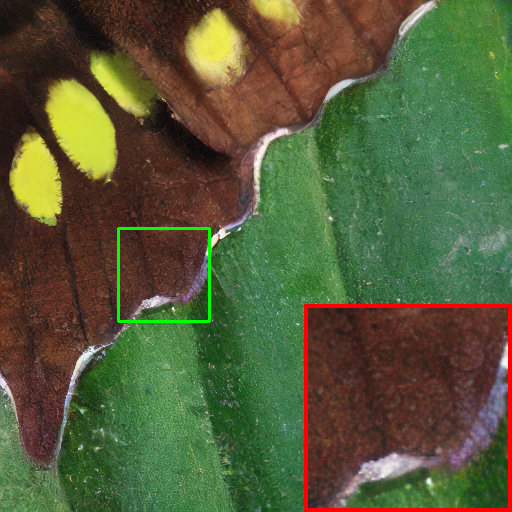} 
        \caption*{SinSR}
    \end{subfigure}
    \hfill
    \begin{subfigure}[b]{0.16\textwidth}
        \centering
        \includegraphics[width=\textwidth]{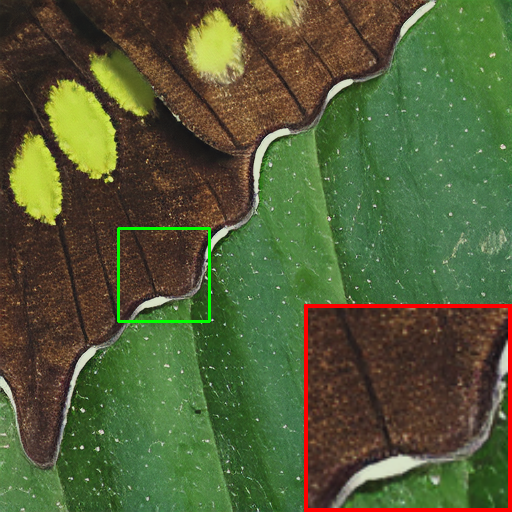} 
        \caption*{Ours}
    \end{subfigure}
    \vspace{0.5em} % Add some vertical space between rows

    % Row 5
    \begin{subfigure}[b]{0.16\textwidth}
        \centering
        \includegraphics[width=\textwidth]{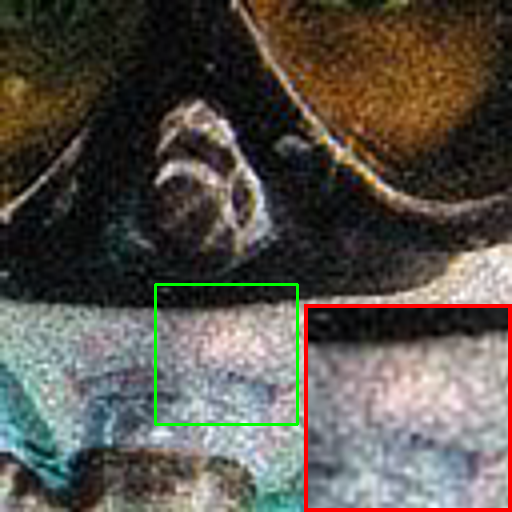} 
        \caption*{Low-Quality}
    \end{subfigure}
    \hfill
    \begin{subfigure}[b]{0.16\textwidth}
        \centering
        \includegraphics[width=\textwidth]{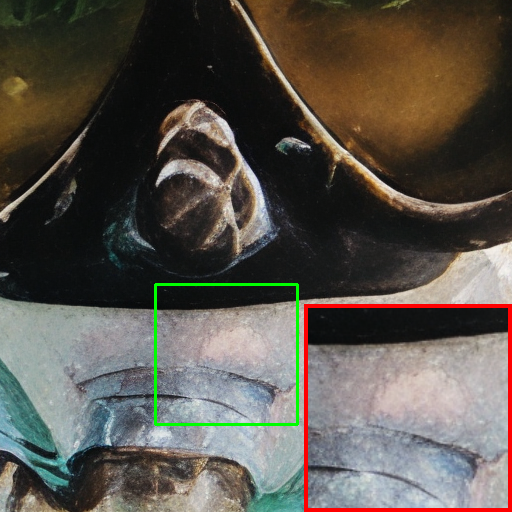} 
        \caption*{OSEDiff}
    \end{subfigure}
    \hfill
    \begin{subfigure}[b]{0.16\textwidth}
        \centering
        \includegraphics[width=\textwidth]{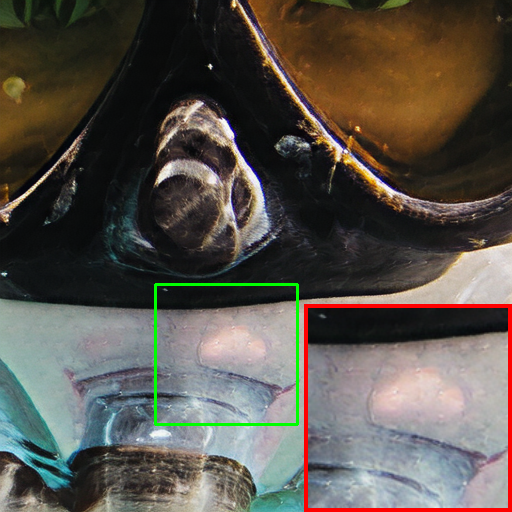}
        \caption*{TSD-SR}
    \end{subfigure}
    \hfill
    \begin{subfigure}[b]{0.16\textwidth}
        \centering
        \includegraphics[width=\textwidth]{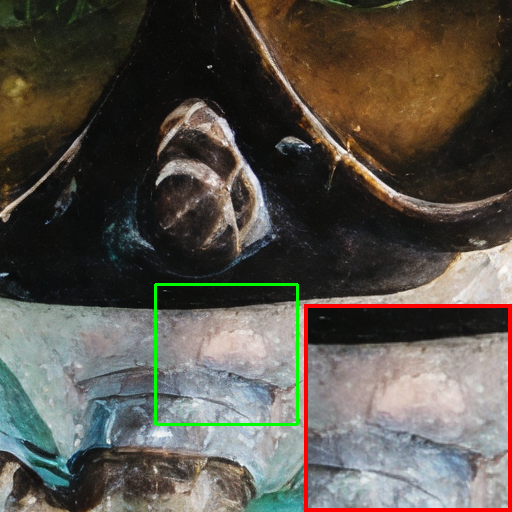} 
        \caption*{AdcSR}
    \end{subfigure}
    \hfill
    \begin{subfigure}[b]{0.16\textwidth}
        \centering
        \includegraphics[width=\textwidth]{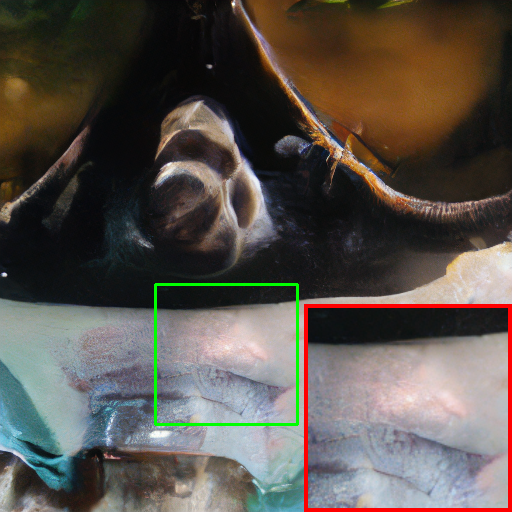} 
        \caption*{SinSR}
    \end{subfigure}
    \hfill
    \begin{subfigure}[b]{0.16\textwidth}
        \centering
        \includegraphics[width=\textwidth]{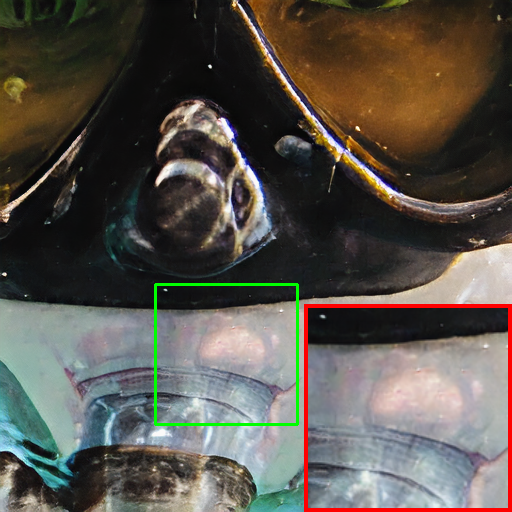} 
        \caption*{Ours}
    \end{subfigure}
    \vspace{0.5em} % Add some vertical space between rows

    % Row 6
    \begin{subfigure}[b]{0.16\textwidth}
        \centering
        \includegraphics[width=\textwidth]{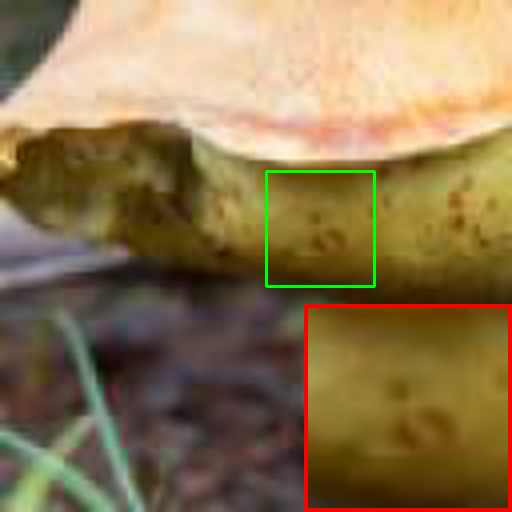} 
        \caption*{Low-Quality}
    \end{subfigure}
    \hfill
    \begin{subfigure}[b]{0.16\textwidth}
        \centering
        \includegraphics[width=\textwidth]{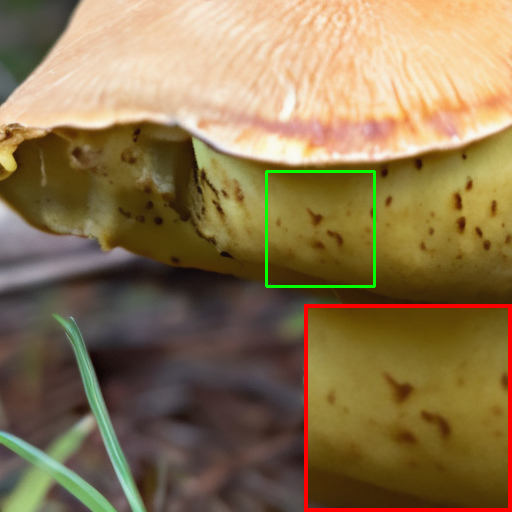} 
        \caption*{SeeSR}
    \end{subfigure}
    \hfill
    \begin{subfigure}[b]{0.16\textwidth}
        \centering
        \includegraphics[width=\textwidth]{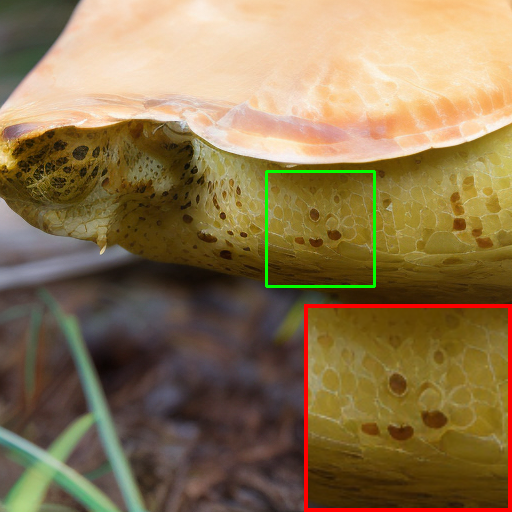} 
        \caption*{DiffBIR}
    \end{subfigure}
    \hfill
    \begin{subfigure}[b]{0.16\textwidth}
        \centering
        \includegraphics[width=\textwidth]{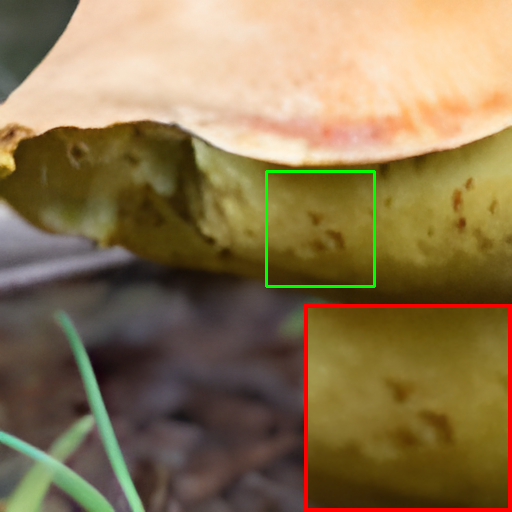} 
        \caption*{Real-ESRGAN}
    \end{subfigure}
    \hfill
    \begin{subfigure}[b]{0.16\textwidth}
        \centering
        \includegraphics[width=\textwidth]{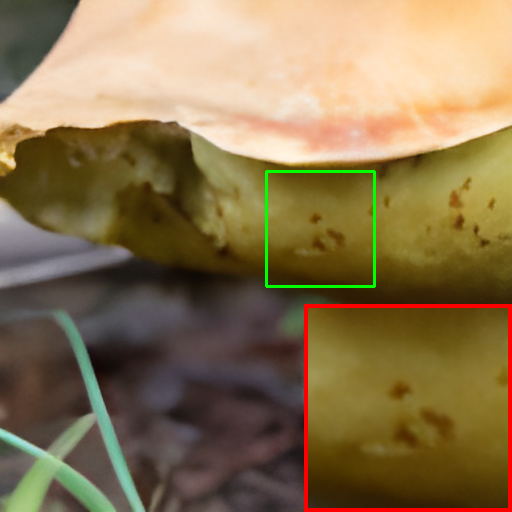} 
        \caption*{FeMASR}
    \end{subfigure}
    \hfill
    \begin{subfigure}[b]{0.16\textwidth}
        \centering
        \includegraphics[width=\textwidth]{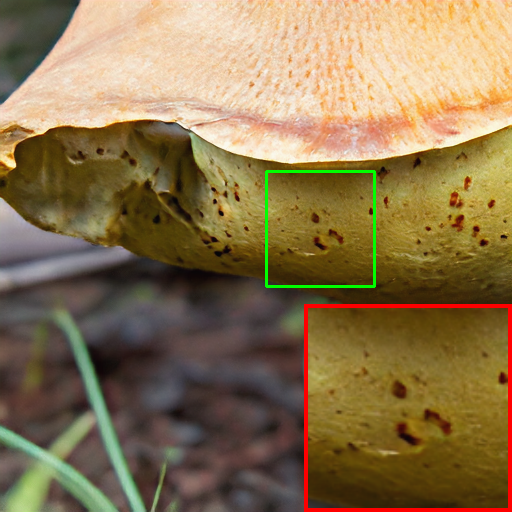} 
        \caption*{Ours}
    \end{subfigure}
    \caption{Qualitative comparisons of GAN-based and diffusion-based Real-ISR methods. 
    Please zoom in for a better view.}
    \label{fig:appendix_quality}
\end{figure*}

\subsection{More Qualitative  Comparisons}
\Cref{fig:appendix_quality} presents a visual comparison between the GAN-based and diffusion-based methods. 
GAN-based methods often struggle to recover fine, high-frequency details, resulting in blurred textures. For instance, models such as BSRGAN, Real-ESRGAN, LDL, and FeMASR produce blurring on petal textures. Similarly, BSRGAN and LDL create overly smooth butterfly wings, while Real-ESRGAN and FeMASR fail to reconstruct crisp mushroom textures. 
This consistent lack of detail suggests a fundamental limitation in the ability of these GAN-based approaches to restore high-frequency information.
Multi-step diffusion-based methods, such as StableSR, DiffBIR, SeeSR, and ResShift, can introduce artifacts when restoring natural textures like water and rocks, and may also produce blurred details. Notably, DiffBIR is particularly susceptible to over-generation, which can result in illogical or unnatural textures, as has been observed in the restoration of images containing mushrooms.
Methods like OSEdiff, AdcSR, and SinSR can suffer from incomplete denoising and are prone to generating broken or fragmented textures during the super resolution process.
Our model demonstrates highly competitive performance, excelling in both structural and texture recovery. Compared to other methods, it restores a greater degree of high-frequency detail while rigorously maintaining overall structural integrity.

\section{More Ablation Studies.}

\subsection{Ablation Study on Prompt Condition}
\Cref{tab:aptoken} presents the results of token pruning of the teacher model. 
We found that pruning prompt information does not negatively impact certain full-referenced metrics. 
In fact, some metrics, such as LPIPS and DISTS, even show improvement at specific pruning rates.
Token pruning primarily affects no-referenced metrics. However, we observe no significant performance degradation even at a 50\% pruning ratio. 
Furthermore, performance degrades gracefully at higher pruning percentages without a sharp decline, which suggests that the contribution of textual information to the final image synthesis is limited.
As shown in \Cref{fig:aptoken-pruning}, although TP 90\% token's output contains less fine-grained detail than the baseline, it effectively removes the noise from the low-quality input, resulting in an image with high visual quality.

\begin{figure}[!t]
    \centering
    % Row 1
    \begin{subfigure}[b]{0.15\textwidth}
        \centering
        \includegraphics[width=\textwidth]{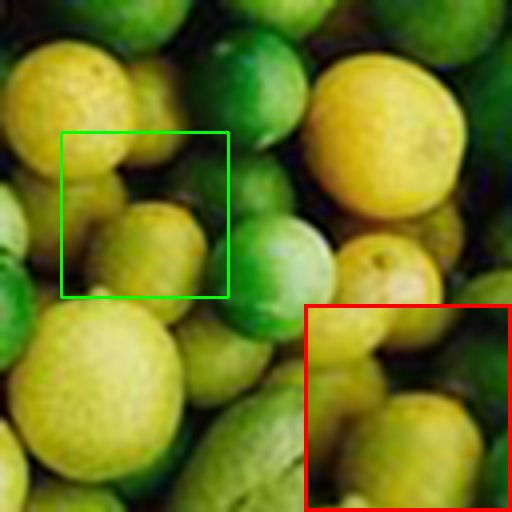} 
        % \caption{(a) Low-Quality}
        % \label{fig:lr_token}
    \end{subfigure}
    \hfill
    \begin{subfigure}[b]{0.15\textwidth}
        \centering
        \includegraphics[width=\textwidth]{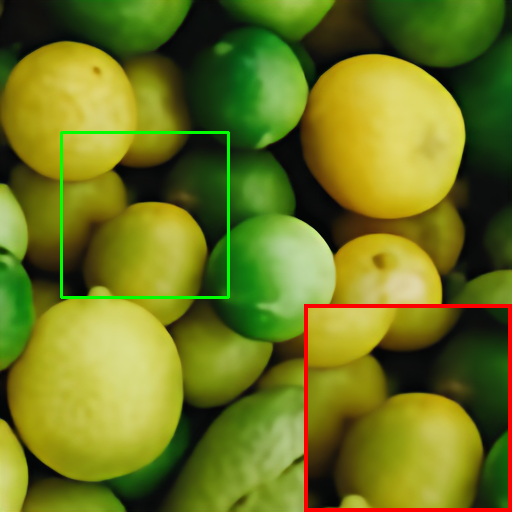} 
        % \caption{(b) Baseline}
        % \label{fig:tsd_token}
    \end{subfigure}
    \hfill
    \begin{subfigure}[b]{0.15\textwidth}
        \centering
        \includegraphics[width=\textwidth]{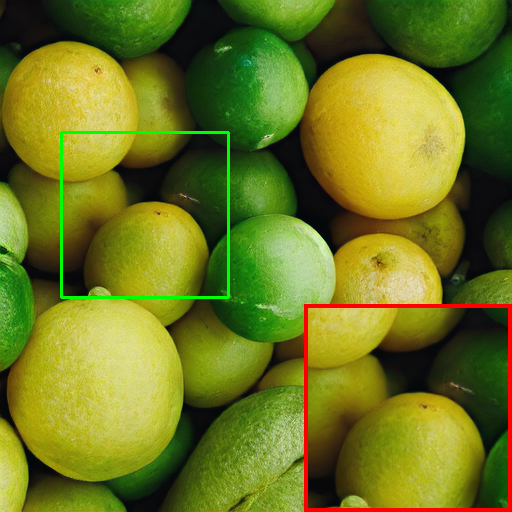} 
        % \caption{(c) 90\% TP Result}
        % \label{fig:out_token}
    \end{subfigure}
    
    \vspace{0.5em}
    
     \begin{subfigure}[b]{0.15\textwidth}
        \centering
        \includegraphics[width=\textwidth]{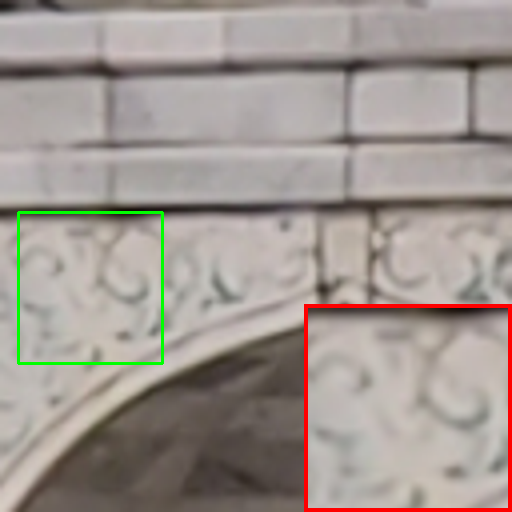} 
        \caption{Low-Quality}
    \end{subfigure}
    \hfill
    \begin{subfigure}[b]{0.15\textwidth}
        \centering
        \includegraphics[width=\textwidth]{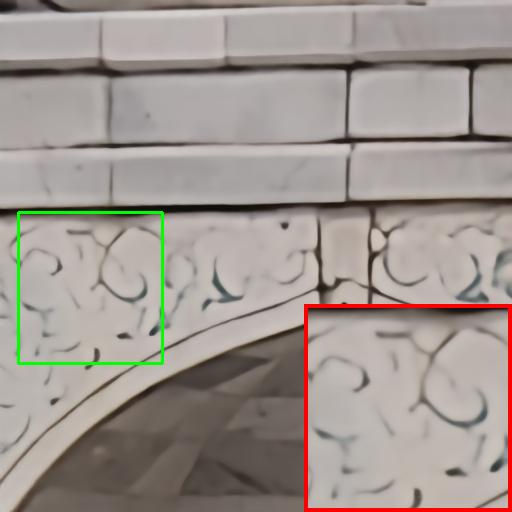} 
        \caption{Distill HR}
    \end{subfigure}
    \hfill
    \begin{subfigure}[b]{0.15\textwidth}
        \centering
        \includegraphics[width=\textwidth]{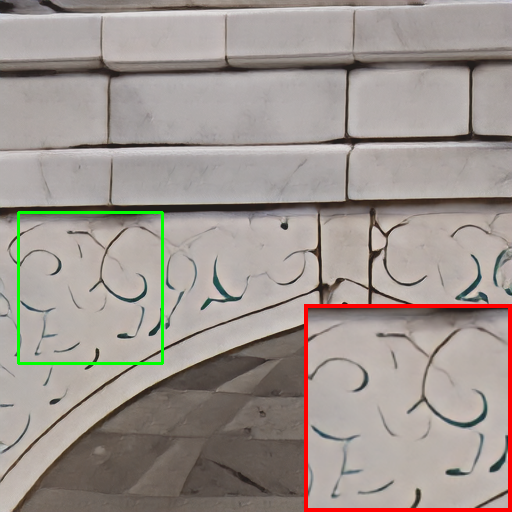} 
        \caption{Distill Tea.}
    \end{subfigure}
    \caption{Visual comparison of knowledge distillation: high-resolution ground truth versus teacher.}
    \label{fig:latent-hr}
    \vspace{-0.1em} % Add some vertical space between rows
\end{figure}

\subsection{Ablation Study on Pruning Ratio}
\Cref{tab:pruning_scheme} compares the performance of our method against ShortGPT and TinyFusion across various metrics at token pruning ratios of 33\%, 50\%, and 67\%.
The results show our approach surpassing TinyFusion \cite{fang2024tinyfusion} and ShotGPT \cite{men2024shortgpt} at every pruning ratio, which demonstrates its robust ability to recover performance.
Furthermore, the model exhibits only a slight degradation in performance as the pruning ratio is increased from 33\% to 50\%, indicating the continued presence of parameter redundancy at the 33\% level.
However, as the pruning rate increases from 50\% to 67\%, the model's performance on metrics such as DISTS and MANIQA declines sharply, indicating that excessive pruning leads to irreversible performance degradation.

\begin{table}[!t]
\centering
\caption{Ablation study of pruning ratio on DIV2K-Val dataset. The best is highlighted in \textbf{bold}.}
\label{tab:pruning_scheme}
\resizebox{\linewidth}{!}{ 
\begin{tabular}{ll|cccccc}
\toprule
\textbf{Method} & \textbf{Pruning Ratio} &
\multicolumn{1}{c}{\textbf{LPIPS} $\downarrow$} & 
\multicolumn{1}{c}{\textbf{DISTS} $\downarrow$} & 
\multicolumn{1}{c}{\textbf{NIQE} $\downarrow$} & 
\multicolumn{1}{c}{\textbf{MANIQA} $\uparrow$} & 
\multicolumn{1}{c}{\textbf{CLIPIQA} $\uparrow$} \\
\midrule

ShortGPT & \quad \quad 33\%  
 & 0.3049 & 0.2150 & 5.1010 & 0.5727 & 0.7248 \\
TinyFusion & \quad \quad 33\%  
& 0.2808 & 0.1928 & 4.3013  & 0.5912 & 0.7274 \\
\textbf{Ours} & \quad \quad 33\%   
 & \textbf{0.2789} & \textbf{0.1917}  & \textbf{4.1396} & \textbf{0.6071} & \textbf{0.7284} \\
 \midrule
ShortGPT & \quad \quad 50\%  
& 0.2892 & 0.2034 & 4.8874  & 0.5681 & 0.7116 \\
TinyFusion & \quad \quad 50\%  
& 0.2799 & 0.1904 & 4.1705 & 0.5880 & 0.6995 \\
\textbf{Ours} & \quad \quad 50\%  
& \textbf{0.2793} & \textbf{0.1883} & \textbf{4.1500} & \textbf{0.6083} & \textbf{0.7201} \\
\midrule
ShortGPT & \quad \quad 67\%  
& 0.3466 & 0.2476 & 5.5182 & 0.5257 & 0.7069 \\
TinyFusion & \quad \quad 67\%  
& 0.3071 & 0.2194 & 4.7256 & 0.5217 & 0.7056 \\
\textbf{Ours} & \quad \quad 67\%  
& \textbf{0.2984} & \textbf{0.2110} & \textbf{4.2475} & \textbf{0.5389} & \textbf{0.7198} \\
\bottomrule
\end{tabular}
}
\end{table}

\begin{table*}[!ht]
\centering
\caption{Ablation studies of Stage 1 distillation loss on DrealSR dataset. The best (\textbf{other than Teacher}) is  highlighted in \textbf{bold}.}
\label{tab:loss}
\resizebox{\linewidth}{!}{ 
\begin{tabular}{l|ccccccccc}
\toprule
\textbf{Method} & 
\multicolumn{1}{c}{\textbf{SSIM} $\uparrow$} & 
\multicolumn{1}{c}{\textbf{LPIPS} $\downarrow$} & 
\multicolumn{1}{c}{\textbf{DISTS} $\downarrow$} & 
\multicolumn{1}{c}{\textbf{FID} $\downarrow$} & 
\multicolumn{1}{c}{\textbf{MUSIQ} $\uparrow$} & 
\multicolumn{1}{c}{\textbf{CLIPIQA} $\uparrow$} & 
\multicolumn{1}{c}{\textbf{TOPIQ} $\uparrow$} & 
\multicolumn{1}{c}{\textbf{Q-Align} $\uparrow$} \\
\midrule
Teacher TSD-SR Baseline 
 & 0.7559 & 0.2967 & {0.2136} & {134.98} & {66.62} & {0.7344} & {0.6177} & 3.6055  \\
 Distill HR in Image Space   
 & \textbf{0.8480} & 0.3082 & 0.2438 & 176.50 & 48.89 & 0.3456 & 0.3652 & 2.4315 \\
Distill TEA. in Image Space 
 & {0.7904}& \textbf{0.2819} & \textbf{0.2150} & 154.16 & 64.74 & 0.6171 & 0.6020 & 3.2871 \\
Distill HR in Latent Space 
 & 0.7814 & 0.4253 & 0.2988 & 190.11 & 48.41  & 0.4441 & 0.4726 & 2.3841 \\
\textbf{Distill TEA. in Latent Space (Ours)}         
& 0.7508    & 0.3316    & 0.2322    &  \textbf{148.63 }  &  \textbf{66.57}   & \textbf{0.7321}   &  \textbf{0.6211}  & \textbf{3.5356}     \\
\bottomrule
\end{tabular}
}
\end{table*}

\begin{table}[!t]
\centering
\caption{Ablation studies of Stage 2 training loss on RealSR dataset. The best is highlighted in \textbf{bold}.}
\label{tab:stage2}
\resizebox{\linewidth}{!}{ 
\begin{tabular}{l|ccccc}
\toprule
\textbf{Method} & 
\multicolumn{1}{c}{\textbf{LPIPS} $\downarrow$} & 
\multicolumn{1}{c}{\textbf{DISTS} $\downarrow$} & 
\multicolumn{1}{c}{\textbf{NIQE} $\downarrow$} & 
\multicolumn{1}{c}{\textbf{MANIQA} $\uparrow$} & 
\multicolumn{1}{c}{\textbf{FID} $\downarrow$} \\
\midrule
Stage 1 Baseline         
& 0.3087    & 0.2302    &  5.1740   & 0.6045   &  132.53     \\
w/ LPIPS loss \& w/o GAN 
& \textbf{0.2702 } & 0.2180 & 5.0696 & 0.5850 & 123,43 \\
w/ GAN loss \& w/o LPIPS
& 0.2844 & 0.2173 & \textbf{4.7203} & 0.6073  & 124.22 \\
\textbf{w/ LPIPS \& w/ GAN (Ours)}   & 0.2806 & \textbf{0.2123} & 4.7400 & \textbf{0.6235}  & \textbf{118.01} \\
\bottomrule
\end{tabular}
}
\end{table}

\subsection{Ablation Study of Knowledge Distillation}
\Cref{tab:loss} demonstrates the effectiveness of our knowledge distillation method under stage 1.
We can draw the following conclusions: (1) Distillation employing GT (High-Quality) data consistently resulted in unsatisfactory performance, whether applied in the image space or the latent space. 
As illustrated in \Cref{fig:latent-hr}, distillation using GT data yields smooth, blurred results, whereas using the teacher produces clearer textures.
(2) Distillation performed in the image space achieves better scores on full-reference metrics such as SSIM, LPIPS, and DISTS.
Distillation in the latent space yields superior no-reference metrics (MUSIQ, CLIPIQA, TOPIQ and Q-Align), with most even matching those of the teacher model. However, a potential compromise in reference metrics necessitates a second stage of training, which we perform in the image space.

\subsection{Ablation Study of Losses in Stage 2}
We conduct an ablation study on the Stage 2 training losses, as shown in \Cref{tab:stage2}. The results indicate that Stage 2 training significantly improved image quality, particularly in terms of image fidelity.
Specifically, we find that the inclusion of LPIPS loss is highly beneficial for improving reference metrics such as DISTS and FID. The addition of GAN loss, in turn, is helpful for enhancing several no-reference metrics, including NIQE and MANIQA.
We ultimately weighted the two new losses to balance the trade-off between fidelity and the generative ability.

% \input{sec/6_appendix}

\end{document}